\documentclass{article}
\usepackage{iclr2027_conference,times}
\usepackage[T1]{fontenc}
\usepackage{graphicx}
\usepackage{helvet}
\usepackage{courier}
\usepackage{amsmath}
\usepackage{algorithm}
\usepackage{algorithmic}
\usepackage{csquotes} 
\usepackage{color}
\usepackage{paralist}
\usepackage{amssymb}
\usepackage{subfigure}
\usepackage{float}
\usepackage{multirow}
\usepackage{mathrsfs}

\usepackage{mathrsfs} 
\usepackage{amsfonts}
\usepackage{pgfplots} 
\usepackage{epstopdf}
\usepackage{epsfig}
\pgfplotsset{compat=newest}
\usepackage{color}
\definecolor{forestgreen}{RGB}{0,139,69}

\usepackage{xcolor}
\definecolor{citecolor}{HTML}{0071bc}
\usepackage[hidelinks]{hyperref}

\usepackage{xcolor}
\definecolor{SeaGreen4}{RGB}{0,205,102} 
\definecolor{SlateBlue}{RGB}{106,90,205} 
\definecolor{DarkRed}{RGB}{178,34,34} 

\usepackage{textcomp,booktabs}
\usepackage{amssymb}
\usepackage{pifont}
\newcommand{\cmark}{\ding{51}}%
\newcommand{\xmark}{\ding{55}}%

\usepackage{makecell}

\usepackage{colortbl}
\definecolor{mygray}{gray}{.9}
\definecolor{mypink}{rgb}{.99,.91,.95}
\definecolor{mycyan}{cmyk}{.3,0,0,0}

\title{SpectralTF-MAE: A Pre-trained Spectral Foundation Model for Tokamak Fusion Plasma Diagnostics}

\author{
\begin{minipage}[t]{0.96\textwidth}
\normalfont\raggedright
{\small\bfseries
Xiao Wang\textsuperscript{1}, Zhijie Lu\textsuperscript{1}, Duoqing Yang\textsuperscript{1}, Jiangtao Yu\textsuperscript{1}, Jin Tang\textsuperscript{1}, Zichao Lin\textsuperscript{2,*}, Hongming Zhang\textsuperscript{2}, Guosheng Xu\textsuperscript{2}, Bo Lyu\textsuperscript{2}, Qiang Liu\textsuperscript{3}}\\[5pt]
{\footnotesize
1. School of Computer Science and Technology, Anhui University, Hefei 230601, China\\
2. Institute of Plasma Physics, Hefei Institutes of Physical Science, Chinese Academy of Sciences, Hefei 230031, China\\
3. New Laboratory of Pattern Recognition, Institute of Automation, Chinese Academy of Sciences, Beijing, China\\
\texttt{\{xiaowang, tangjin\}@ahu.edu.cn}\\
\texttt{\{zichao.lin, hmzhang, blu\}@ipp.ac.cn}\\
\texttt{qiang.liu@nlpr.ia.ac.cn}\\
\textit{Corresponding author: Zichao Lin}
}
\end{minipage}
}

\iclrfinalcopy

\begin{document}

\maketitle
\fancyhead[L]{\fontsize{8.5}{10}\selectfont SpectralTF-MAE: A Pre-trained Spectral Foundation Model for Tokamak Fusion Plasma Diagnostics}

\begin{abstract}
Spectroscopic diagnostics provide rich information for characterizing plasma states in tokamak fusion experiments, while existing learning-based approaches are often designed for individual tasks and rely heavily on labeled data. In this work, we propose SpectralTF-MAE, a self-supervised spectral foundation model for tokamak plasma diagnostics. The model is pretrained on paired charge-exchange recombination spectroscopy (CXRS) and X-ray crystal spectroscopy (XCS) measurements through masked spectral reconstruction. To better capture diagnostically important spectral structures, we introduce a peak-aware masking strategy together with a peak-enhanced reconstruction objective, encouraging the model to focus on characteristic spectral regions. A shared Transformer encoder is employed to learn transferable representations across different diagnostic modalities. The pretrained model is further adapted to several downstream tasks, including plasma-parameter prediction, cross-diagnostic spectral generation, and spectral peak analysis. Experiments on EAST diagnostic data demonstrate that the proposed framework achieves competitive performance across different tasks and consistently benefits from self-supervised spectral pretraining. These results indicate the potential of SpectralTF-MAE as a general and transferable representation learning framework for plasma spectroscopic diagnostics. The source code and pre-trained models will be released on \url{https://github.com/Event-AHU/OpenFusion/tree/main/SpectralTF-MAE}. 
\end{abstract}

\section{Introduction}

\begin{figure}[!t]
\centering
\setlength{\subfigcapskip}{0pt}

\begin{minipage}[t]{0.49\textwidth}
    \vspace{0pt}
    \centering
    \subfigure[CXRS acquisition.\label{fig:cxrs_acquisition}]{
        \includegraphics[width=0.965\linewidth]{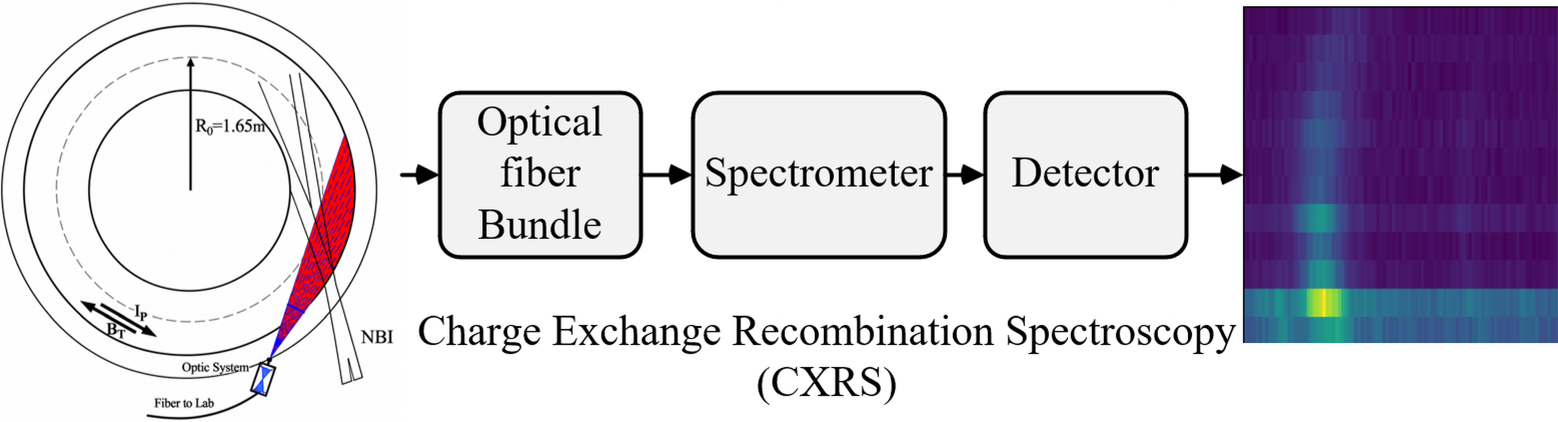}
    }
\end{minipage}
\hfill
\begin{minipage}[t]{0.49\textwidth}
    \vspace{0pt}
    \centering
    \subfigure[XCS acquisition.\label{fig:xcs_acquisition}]{
        \includegraphics[width=0.965\linewidth]{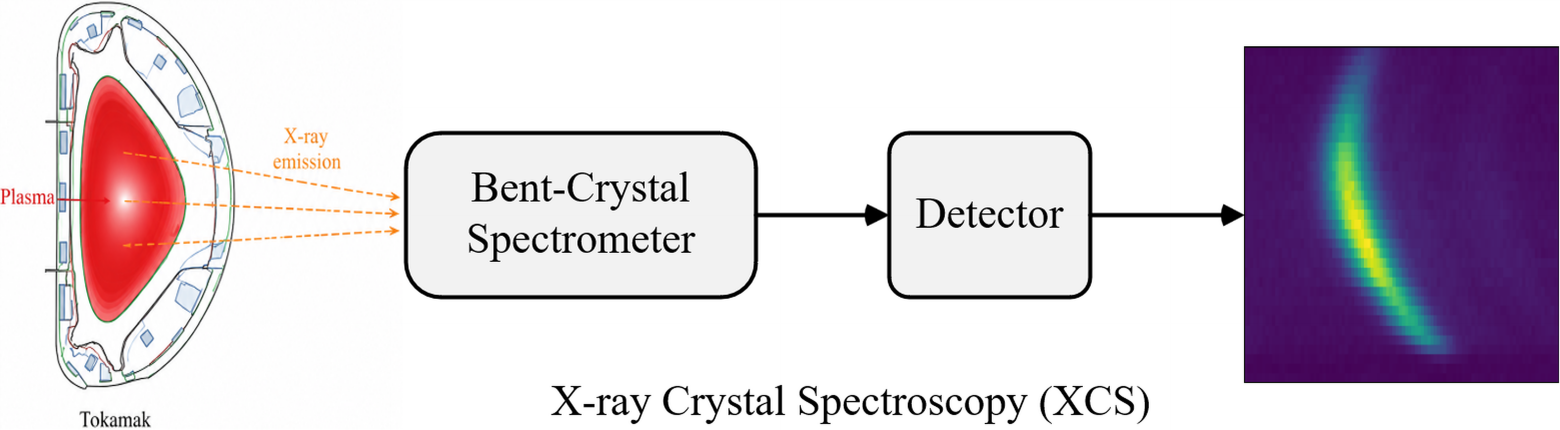}
    }
\end{minipage}

\vspace{0.15em}

\setlength{\subfigcapskip}{-1.2pt}

\begin{minipage}[t]{0.32\textwidth}
    \vspace{0pt}
    \centering
    \subfigure[XCS$\rightarrow$CXRS generation.\label{fig:xcs_to_cxrs_generation}]{%
        \makebox[0.98\linewidth][c]{\includegraphics[height=2.78cm]{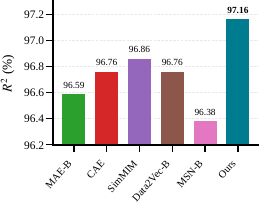}}
    }%
\end{minipage}
\hfill
\begin{minipage}[t]{0.32\textwidth}
    \vspace{0pt}
    \centering
    \subfigure[CXRS$\rightarrow$XCS generation.\label{fig:cxrs_to_xcs_generation}]{%
        \makebox[0.98\linewidth][c]{\includegraphics[height=2.78cm]{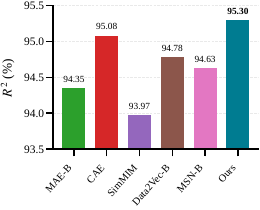}}
    }%
\end{minipage}
\hfill
\begin{minipage}[t]{0.32\textwidth}
    \vspace{0pt}
    \centering
    \subfigure[Spectral peak analysis.\label{fig:peak_analysis_overview}]{%
        \makebox[0.98\linewidth][c]{\includegraphics[height=2.78cm]{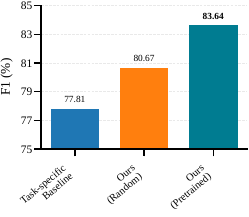}}
    }%
\end{minipage}

\vspace{0.10em}

\begin{minipage}[t]{\textwidth}
    \vspace{0pt}
    \centering
    \subfigure[Plasma-parameter prediction.\label{fig:parameter_prediction_overview}]{
        \includegraphics[width=0.985\linewidth]{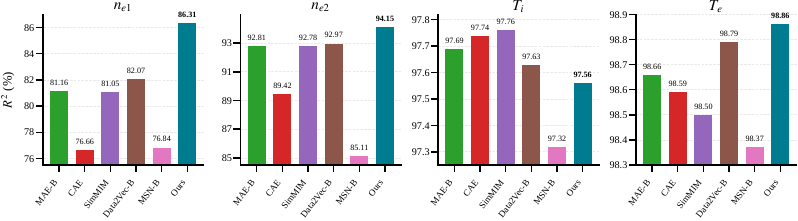}
    }
\end{minipage}

\caption{
(a-b). Demonstration of different principles for spectral‑data acquisition; 
(c-f). Performance comparison of downstream tasks based on our newly proposed spectral foundation model.
}
\label{fig:spectral_acquisition}
\end{figure}

Artificial Intelligence (AI) empowered controlled nuclear fusion has attracted growing attention from researchers~\citep{humphreys2020advancing,degrave2022magnetic,seo2024avoiding}. Spectroscopic diagnosis constitutes one of the core approaches for plasma analysis. On the EAST large-science facility, commonly used spectroscopic diagnostic systems acquire two major categories of spectra: active spectra and passive spectra. Specifically, charge-exchange recombination spectroscopy (CXRS) is an active diagnostic technique that employs external neutral beams to induce charge-exchange reactions and obtain plasma parameters such as ion temperature and rotation velocity, as illustrated in Fig.~\ref{fig:spectral_acquisition}~(a). In contrast, X-ray crystal spectroscopy (XCS) is a passive diagnostic technique that measures intrinsic X-ray emission from plasma and provides information related to electron temperature, impurity behavior, and plasma conditions, as shown in Fig.~\ref{fig:spectral_acquisition}~(b). Existing studies generally adopt small-scale models for spectral interpretation~\citep{chai2019fast,lin2024prediction,lin2025development,saura2025machine,li2025prediction}, which evidently fail to fully exploit the potential of large-scale spectral datasets. Spectral perception tailored for controlled-nuclear-fusion tasks remains an open and pressing challenge.

With the proposal of the Transformer~\citep{vaswani2017attention} and the rapid advancement of large‑model technologies~\citep{DBLP:journals/corr/abs-2108-07258}, a variety of large models for text, vision and even multimodal data have emerged~\citep{devlin-etal-2019-bert,dosovitskiy2020image,radford2021clip}. Leveraging self‑supervised learning, these models capture inherent patterns from massive training data and achieve substantial performance improvements across downstream tasks. Several existing studies~\citep{su2026spectrallm,liu2025molspectra} have investigated large-model-based learning strategies for spectral data, such as SpectraLLM~\citep{su2026spectrallm} and MolSpectra~\citep{liu2025molspectra}. Nevertheless, these approaches have not performed dedicated pre‑training on controlled‑fusion spectral data. Consequently, there remains a lack of domain‑specific large models tailored for this type of data.

Through a systematic investigation of the spectral applications in controlled nuclear fusion research, we summarize three core requirements for plasma spectral analysis. 
\textbf{First}, accurate predictions of plasma temperature (\(T_i\)/\(T_e\)) profiles and electron density (\(n_e\)) from measured spectral data are essential. 
\textbf{Second}, active and passive spectra show inconsistent correspondence, and the prevalent missing data in active spectra severely limits downstream diagnostic performance. 
\textbf{Third}, quantitative analysis of spectral peak intensity and spectral broadening is required to provide reliable data support for subsequent plasma parameter inversion and follow-up research tasks. 
Existing studies~\citep{chai2019fast,samuell2021measuring,lin2024prediction,lin2025development,lee2026neural,saura2025machine,li2025prediction} typically design independent network architectures for individual sub-tasks and train corresponding surrogate models using small-scale networks. Such paradigms fail to fully explore the potential of massive spectral data and cannot leverage the powerful data fitting and feature mining capabilities of large models.

\begin{figure}
\centering
\includegraphics[width=1\linewidth]{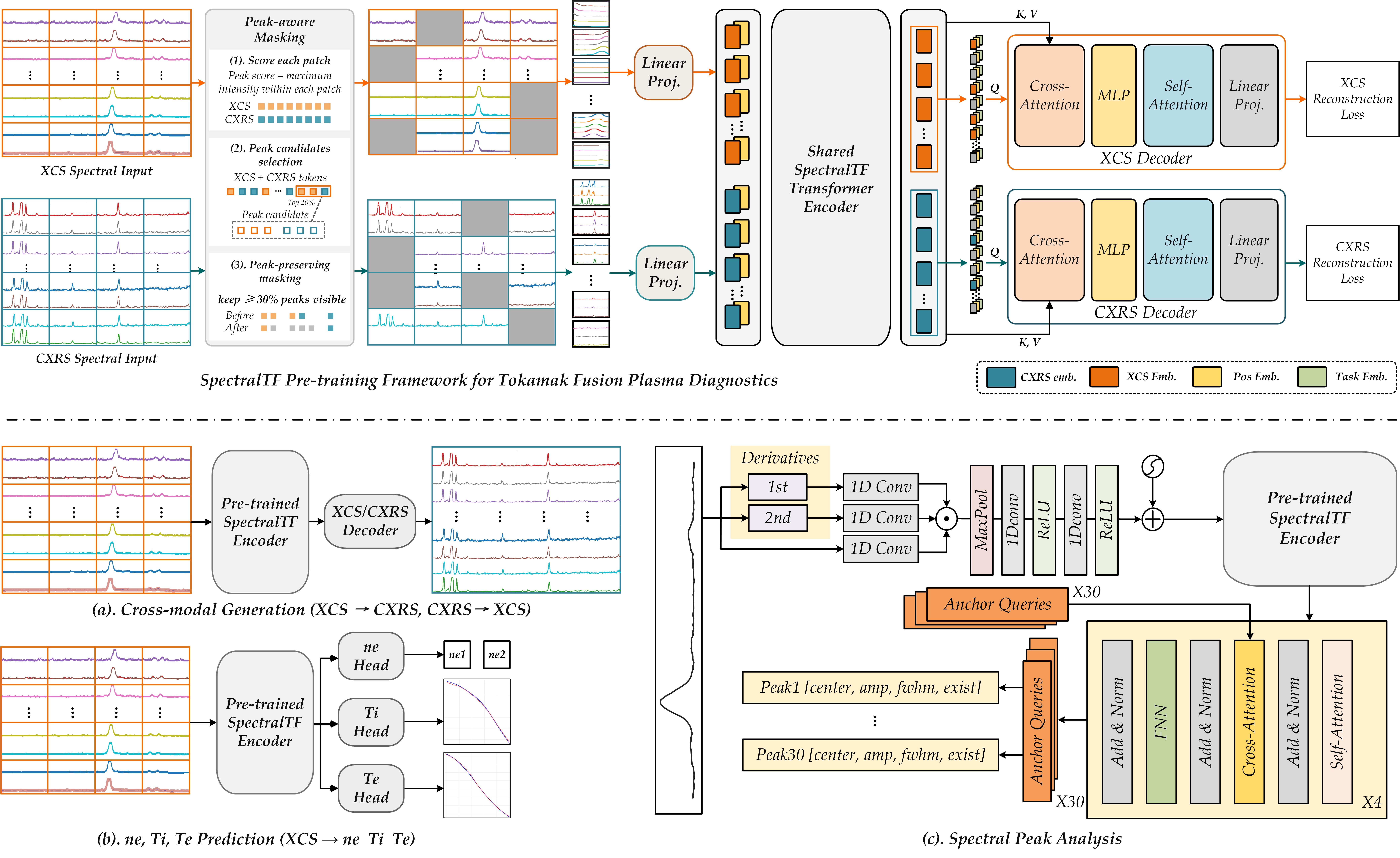}
\caption{An overview of our proposed pre-trained spectral foundation model (upper panel) and downstream tasks (lower panel) for tokamak fusion plasma diagnostics. 
}
\label{fig:framework}
\end{figure}

In this paper, we propose a novel pre-trained spectral foundation model for tokamak fusion plasma diagnostics based on Masked Auto-Encoder (MAE)~\citep{He_2022_CVPR}, termed SpectralTF-MAE. As shown in Fig.~\ref{fig:framework}, given the XCS and CXRS spectral input, we first perform spectral‑peak detection and adaptive masking on it. Different from the random masking strategy adopted in vanilla MAE, our adaptive masking implements automatic thresholding over high‑amplitude spectral‑peak regions. This physics‑aware masking operation prioritizes the most informative peak components for corruption, which forces the model to learn underlying physical correlations of fusion spectra rather than trivial background patterns. Then, we project the remaining spectral patches into tokens using a linear layer and extract the deep representations using a shard spectral Transformer encoder network. After that, the XCS and CXRS decoder networks are utilized for masked spectral reconstruction. We conduct experiments on three tasks to validate the effectiveness of our pre-trained spectral foundation model SpectralTF-MAE, including cross-spectral generation (i.e., XCS $\rightarrow$ CXRS, CXRS $\rightarrow$ XCS), \(T_i\)/\(T_e\) profile prediction, electron-density (\(n_e\)) prediction, and spectral peak analysis.

To sum up, the main contributions of this paper can be summarized as follows: 

1). We propose a novel foundation model for controllable‑fusion spectral data, termed SpectralTF-MAE, which builds on the Masked Auto‑Encoder (MAE) self‑supervised learning framework and adopts a physics‑aware masking strategy to incentivize the model to learn intrinsic plasma spectral characteristics. 

2). Based on the proposed spectral foundation model, we develop a suite of downstream tasks covering spectral generation and analysis, including cross‑spectral generation, \(T_i\)/\(T_e\) profile prediction, electron-density (\(n_e\)) prediction, and spectral‑peak analysis, which establishes a solid model benchmark for large model driven controlled nuclear fusion spectral analysis. 

3). Extensive experiments with comprehensive quantitative and qualitative analysis on the newly collected spectral dataset fully validated the effectiveness and generalization of our proposed pre-trained spectral foundation model.

\section{Related Works}

Foundation models learn transferable representations through large-scale pretraining and downstream adaptation. Transformer-based masked or predictive self-supervised learning has been successful in language and vision~\citep{vaswani2017attention,devlin-etal-2019-bert,He_2022_CVPR,bao2021beit,xie2022simmim,chen2024context,baevski2022data2vec}, and related ideas have been extended to hyperspectral and other scientific spectral data~\citep{hong2021spectralformer,cong2022satmae,10490262,wang2025hypersigma,parker2024astroclip,leung2024towards}. These studies motivate reusable spectral representations, but plasma spectroscopy remains comparatively underexplored.

Machine learning has also been widely investigated for magnetic-confinement fusion, including disruption prediction, diagnostic inference, and plasma control~\citep{humphreys2020advancing,rea2018disruption,montes2019machine,degrave2022magnetic,seo2024avoiding}. More recently, reusable pretrained models have begun to appear in fusion research~\citep{churchill2025ai,yang2026fusionmae}. Existing fusion foundation-model studies primarily emphasize heterogeneous diagnostic time series and global plasma-state representations, whereas our work focuses on high-dimensional XCS and CXRS measurements and transferable spectral representations.

Spectroscopic diagnostics provide measurements of plasma temperature, density, impurity states, and rotation~\citep{weller2010diagnostics,mazon2025diagnostics}. Learning-based methods have been developed for temperature and rotation inference from fusion spectra~\citep{chai2019fast,samuell2021measuring,lin2024prediction,lin2025development,lee2026neural}, but these approaches are typically task-specific and fully supervised. In contrast, SpectralTF-MAE learns a shared representation from paired XCS and CXRS measurements and transfers it to plasma-parameter prediction, cross-diagnostic generation, and spectral peak analysis. Additional related-work discussion is provided in Appendix~\ref{app:related_work}.

\section{Our Proposed Approach} 

\subsection{Overview} 

We propose a self-supervised spectral representation learning framework for plasma diagnostics, consisting of multi-diagnostic spectral pretraining and downstream adaptation. As illustrated in Fig.~\ref{fig:framework}, paired charge-exchange recombination spectroscopy (CXRS) and X-ray crystal spectroscopy (XCS) measurements are jointly modeled through masked spectral reconstruction without requiring plasma-parameter labels. The two spectral modalities are embedded into a shared latent space and processed by a Transformer encoder~\citep{vaswani2017attention} to capture both modality-specific structures and cross-diagnostic correlations. To incorporate plasma-specific spectral characteristics, we introduce a peak-aware masked modeling strategy that emphasizes informative emission regions and encourages the preservation of physically relevant spectral features. After pretraining, the learned representation is adapted to downstream tasks, including plasma-parameter inference from XCS spectra and cross-diagnostic spectral generation between CXRS and XCS. This framework enables the utilization of paired unlabeled spectral data to learn transferable representations for diverse plasma diagnostic applications.

\subsection{Pretraining Stage}

The pretraining stage learns a shared spectral representation from paired CXRS and XCS measurements without plasma-parameter labels. As shown in Fig.~\ref{fig:framework}, both spectra are normalized and converted into modality-specific tokens. Peak-aware masking defines the reconstruction targets, while the remaining visible tokens are jointly encoded by a shared Transformer. Modality-specific decoders then reconstruct the two diagnostics from the shared latent representation, and a peak-enhanced objective places additional emphasis on the characteristic spectral regions.

Let $\widetilde{\mathbf{X}}^{m}$ denote the normalized spectrum for modality $m\in\{c,x\}$. Modality-specific rectangular patches are projected into a common latent space with positional embeddings. To emphasize diagnostically informative responses, the maximum normalized intensity within each patch is used as a peak score, and a sample-specific quantile threshold identifies peak candidates. These candidates are preferentially selected as reconstruction targets while part of the peak-related context remains visible.

The visible CXRS and XCS tokens are concatenated and processed by a shared Transformer encoder,
\begin{equation}
    \mathbf{Z}=\mathcal{F}_{\mathrm{enc}}([\mathcal{V}^{c};\mathcal{V}^{x};\mathbf{g}]),
\end{equation}
where $\mathbf{g}$ is a learnable global token. Separate modality-specific decoders reconstruct CXRS and XCS from the shared representation. For each diagnostic, reconstruction is optimized over the masked region $\mathcal{M}^{m}$, with additional weight assigned to the predefined characteristic region $\Omega_m$. The overall objective is
\begin{equation}
    \mathcal{L}_{\mathrm{pre}}
    =\sum_{m\in\{c,x\}}
    \left(\mathcal{L}_{\mathrm{rec}}^{m}+\lambda_{\mathrm{peak}}\mathcal{L}_{\mathrm{peak}}^{m}\right).
\end{equation}
Peak-aware masking determines which spectral structures are preferentially reconstructed, whereas the peak-enhanced loss increases optimization emphasis on characteristic spectral regions. Full tokenization, masking, reconstruction, and loss definitions are given in Appendix~\ref{app:pretraining_details}.

\subsection{Downstream Tasks}

After self-supervised pretraining, the learned representation is adapted to three task families: plasma-parameter prediction from XCS measurements, bidirectional cross-diagnostic spectral generation between XCS and CXRS, and spectral peak analysis. These tasks evaluate physical-parameter inference, cross-diagnostic correspondence, and spectral-component analysis, respectively.

\noindent $\bullet$ \textbf{Plasma-Parameter Prediction.~}
For temperature prediction, $T_i$ and $T_e$ are radial-profile regression targets represented at 32 radial positions. For electron density, two measurements are used: $n_{e1}$ denotes the solid-source measurement and $n_{e2}$ is obtained from POINT. After consistency-based screening, 5,330 valid samples are retained. The complete XCS spectrum is encoded and passed to independent task-specific regression heads,
\begin{equation}
\widehat{\mathbf{y}}_{k}=\mathcal{H}_{k}(\mathbf{f}_{x}),\qquad k\in\{n_{e1},n_{e2},T_i,T_e\}.
\end{equation}

\noindent $\bullet$ \textbf{Cross-Diagnostic Spectral Generation.~}
Given a complete source spectrum, the pretrained encoder and target-diagnostic decoder reconstruct the complete target spectrum,
\begin{equation}
\widehat{\mathbf{X}}^{t}=\mathcal{D}_{t}(\mathcal{F}_{\mathrm{enc}}(\mathbf{X}^{s})),\qquad s\neq t.
\end{equation}
XCS-to-CXRS and CXRS-to-XCS directions are fine-tuned independently, with supervision concentrated on the characteristic peak region of the target diagnostic.

\noindent $\bullet$ \textbf{Spectral Peak Analysis.~}
The downstream peak-analysis model decomposes an input spectrum into multiple components and predicts the center, amplitude, full width at half maximum (FWHM), and existence probability of each peak. To isolate the contribution of pretraining, the original task-specific pipeline is retained and only its Transformer encoder is replaced by the pretrained SpectralTF encoder. Additional task details are provided in Appendix~\ref{app:downstream_details}.

\section{Experiments}

\subsection{Datasets and Evaluation Metrics}
The spectral dataset contains paired CXRS and XCS measurements acquired during EAST experiments from 2021 to 2023. After synchronization and quality control, 15,303 valid paired samples are retained. The complete CXRS input is represented as a $1024\times12$ spectral matrix and XCS is spatially integrated to a $35\times487$ representation. Training-only augmentation jointly shifts paired spectra along the wavelength direction and adds Poisson noise; validation and test samples are not augmented.

Regression and spectral-generation tasks are evaluated using MAE, RMSE, and $R^2$. For $T_i$ and $T_e$, metrics aggregate all test samples and all 32 radial positions. For cross-diagnostic generation, metrics are computed in the normalized characteristic region of the target diagnostic. Spectral peak analysis uses detection, localization, shape, and reconstruction metrics. Detailed preprocessing and metric definitions are provided in Appendix~\ref{app:data_metrics}.

\subsection{Implementation Details}

The framework was implemented in PyTorch~\citep{paszke2019pytorch}, and all experiments were conducted on an NVIDIA GeForce RTX 3090 GPU. SpectralTF-MAE-Base is used as the default configuration throughout the study. SpectralTF-MAE-Large is additionally evaluated on the two main quantitative tasks, namely bidirectional cross-diagnostic spectral generation and plasma-parameter prediction; unless otherwise stated, the remaining analyses and ablation studies use the Base configuration.

All models are optimized using AdamW~\citep{loshchilov2017decoupled}. Self-supervised pretraining is conducted for 800 epochs with a batch size of 128 and an effective learning rate of $5\times10^{-5}$. Plasma-parameter prediction and each cross-diagnostic generation direction are fine-tuned for 200 epochs with a batch size of 32 and an effective learning rate of $6.25\times10^{-6}$. During pretraining, approximately 80\% of the combined CXRS--XCS tokens are masked, whereas complete spectra are used without masking for downstream adaptation. The complete pretrained backbone is fine-tuned rather than frozen. For plasma-parameter prediction, targets are standardized using statistics estimated exclusively from the training set.

Model performance is monitored on the validation set, and the checkpoint achieving the best validation $R^2$-based criterion is retained for evaluation on the held-out test set. Hyperparameters are selected using the validation set only.

\subsection{Comparison With Other Methods}

The baselines include conventional convolutional and Transformer backbones as well as representative self-supervised pretraining methods. ResNet-50 and ResNet-101 follow the residual-learning architecture of He et al.~\citeyearpar{He_2016_CVPR}, and the ViT variants follow the Vision Transformer design~\citep{dosovitskiy2020image}. The self-supervised baselines include MAE~\citep{He_2022_CVPR}, CAE~\citep{chen2024context}, SimMIM~\citep{xie2022simmim}, data2vec~\citep{baevski2022data2vec}, and MSN~\citep{assran2022masked}. To make the model-scale comparison explicit, Base and Large variants are reported for ViT, MAE, data2vec, and MSN where applicable, alongside the Base and Large configurations of SpectralTF-MAE. Each baseline is initialized from a pretrained checkpoint released by its original authors. Downstream adaptation is controlled across methods by using the same task-specific datasets, downstream heads, fine-tuning strategy, and evaluation protocol. This setup preserves each method's published pretrained initialization while holding the downstream training conditions fixed.

\noindent $\bullet$ \textbf{Results on Bidirectional Spectral Generation.~}

Table~\ref{tab:bidirectional_generation} compares the evaluated methods for both XCS-to-CXRS and CXRS-to-XCS spectral generation. The two generation directions are fine-tuned independently using the corresponding source--target pairs.

\begin{table}[t]
\centering
\caption{Comparison of methods on bidirectional cross-diagnostic spectral generation. For XCS-to-CXRS generation, MAE and RMSE are calculated in the normalized CXRS characteristic peak region; for CXRS-to-XCS generation, they are calculated in the normalized XCS characteristic peak region. All $R^2$ values are multiplied by 100.}
\label{tab:bidirectional_generation}
\setlength{\tabcolsep}{4.2pt}
\renewcommand{\arraystretch}{1.15}
\resizebox{\linewidth}{!}{
\begin{tabular}{lccc|ccc}
\toprule
\multirow{2}{*}{Method} &
\multicolumn{3}{c|}{XCS $\rightarrow$ CXRS} &
\multicolumn{3}{c}{CXRS $\rightarrow$ XCS} \\
\cmidrule(lr){2-4}\cmidrule(lr){5-7}
&
MAE $\downarrow$ & RMSE $\downarrow$ & $R^2$ $\uparrow$ &
MAE $\downarrow$ & RMSE $\downarrow$ & $R^2$ $\uparrow$ \\
\midrule

ResNet-50~\citep{He_2016_CVPR}
& 0.00683 & 0.01398 & 96.90
& 0.02866 & 0.04759 & 90.04 \\

ResNet-101~\citep{He_2016_CVPR}
& 0.00677 & 0.01403 & 96.88
& 0.02880 & 0.04912 & 89.39 \\

ViT-Small~\citep{dosovitskiy2020image}
& 0.01540 & 0.02625 & 89.08
& 0.06026 & 0.11271 & 44.13 \\

ViT-Base~\citep{dosovitskiy2020image}
& 0.01552 & 0.02655 & 88.83
& 0.06012 & 0.11262 & 44.21 \\

ViT-Large~\citep{dosovitskiy2020image}
& 0.01522 & 0.02603 & 89.26
& 0.05993 & 0.11254 & 44.29 \\

MAE-Base~\citep{He_2022_CVPR}
& 0.00706 & 0.01466 & 96.59
& 0.02215 & 0.03583 & 94.35 \\

MAE-Large~\citep{He_2022_CVPR}
& 0.00697 & 0.01431 & 96.76
& 0.02171 & 0.03531 & 94.52 \\

CAE~\citep{chen2024context}
& 0.00730 & 0.01429 & 96.76
& 0.01974 & 0.03342 & 95.08 \\

SimMIM~\citep{xie2022simmim}
& 0.00716 & 0.01407 & 96.86
& 0.02306 & 0.03701 & 93.97 \\

Data2Vec-Base~\citep{baevski2022data2vec}
& 0.00705 & 0.01429 & 96.76
& 0.02055 & 0.03444 & 94.78 \\

Data2Vec-Large~\citep{baevski2022data2vec}
& 0.00692 & 0.01429 & 96.77
& 0.01985 & 0.03381 & 94.90 \\

MSN-Base~\citep{assran2022masked}
& 0.00751 & 0.01510 & 96.38
& 0.02088 & 0.03491 & 94.63 \\

MSN-Large~\citep{assran2022masked}
& 0.00698 & 0.01379 & 96.99
& 0.01991 & 0.03351 & 94.99 \\

\midrule

\textbf{SpectralTF-MAE-Base}
& 0.00632 & \textbf{0.01339} & 97.16
& 0.01796 & 0.03269 & 95.30 \\

\textbf{SpectralTF-MAE-Large}
& \textbf{0.00623} & 0.01364 & \textbf{97.30}
& \textbf{0.01760} & \textbf{0.03193} & \textbf{95.45} \\

\bottomrule
\end{tabular}
}
\end{table}

Across the two model scales, SpectralTF-MAE attains the best result for every metric in the bidirectional spectral-generation comparison. For XCS-to-CXRS generation, SpectralTF-MAE-Large achieves the lowest MAE (0.00623) and the highest $R^2$ (97.30), whereas SpectralTF-MAE-Base gives the lowest RMSE (0.01339). This complementary behavior indicates that increasing model capacity improves some aspects of the forward generation task but does not uniformly reduce all error measures. For the reverse CXRS-to-XCS direction, SpectralTF-MAE-Large achieves the best results on all three metrics, with an MAE of 0.01760, an RMSE of 0.03193, and an $R^2$ of 95.45. Overall, the Base and Large variants consistently recover the characteristic target-diagnostic response, while the Large model provides the clearest advantage in the CXRS-to-XCS direction.

\noindent $\bullet$ \textbf{Results on Plasma-Parameter Prediction.~}

Table~\ref{tab:parameter_prediction} reports performance for two electron-density quantities ($n_{e1}$ and $n_{e2}$), ion temperature ($T_i$), and electron temperature ($T_e$). 
\begin{table}[t]
\centering
\caption{Comparison of methods on plasma-parameter prediction from XCS measurements. The $T_i$ and $T_e$ metrics are calculated by aggregating all test samples and all 32 radial positions. All $R^2$ values are multiplied by 100.}
\label{tab:parameter_prediction}
\setlength{\tabcolsep}{2.6pt}
\renewcommand{\arraystretch}{1.15}
\resizebox{\textwidth}{!}{
\begin{tabular}{lccc|ccc|ccc|ccc}
\toprule
\multirow{2}{*}{Method} &
\multicolumn{3}{c|}{$n_{e1}$} &
\multicolumn{3}{c|}{$n_{e2}$} &
\multicolumn{3}{c|}{$T_i$} &
\multicolumn{3}{c}{$T_e$} \\
\cline{2-13}
& MAE $\downarrow$ & RMSE $\downarrow$ & $R^2$ $\uparrow$
& MAE $\downarrow$ & RMSE $\downarrow$ & $R^2$ $\uparrow$
& MAE $\downarrow$ & RMSE $\downarrow$ & $R^2$ $\uparrow$
& MAE $\downarrow$ & RMSE $\downarrow$ & $R^2$ $\uparrow$ \\
\midrule

ResNet-50~\citep{He_2016_CVPR}
& 0.29697 & 0.41943 & 77.18
& 0.16697 & 0.26488 & 89.15
& 44.38147 & 66.45265 & 97.18
& 98.92274 & 165.28667 & 97.86 \\

ResNet-101~\citep{He_2016_CVPR}
& 0.29142 & 0.42682 & 76.37
& 0.18265 & 0.30272 & 85.83
& 48.54834 & 73.08930 & 96.59
& 110.23382 & 185.75727 & 97.30 \\

ViT-Small~\citep{dosovitskiy2020image}
& 0.26455 & 0.37971 & 81.30
& 0.15322 & 0.25030 & 90.31
& 42.46420 & 65.57747 & 97.26
& 86.27301 & 142.69225 & 98.41 \\

ViT-Base~\citep{dosovitskiy2020image}
& 0.24610 & 0.34939 & 84.17
& 0.12798 & 0.20237 & 93.67
& 39.81028 & 61.63069 & 97.58
& 76.46207 & 125.69500 & 98.77 \\

ViT-Large~\citep{dosovitskiy2020image}
& 0.22526 & 0.34130 & 84.89
& 0.12143 & 0.20529 & 93.48
& \textbf{38.71456} & 59.66490 & 97.73
& 73.02069 & 130.84304 & 98.66 \\

MAE-Base~\citep{He_2022_CVPR}
& 0.25405 & 0.38112 & 81.16
& 0.13886 & 0.21567 & 92.81
& 40.36044 & 60.22436 & 97.69
& 78.77065 & 130.91341 & 98.66 \\

MAE-Large~\citep{He_2022_CVPR}
& 0.23208 & 0.34132 & 86.09
& 0.12361 & 0.20091 & 93.75
& 47.79674 & 172.89531 & 88.65
& 81.62728 & 153.79097 & 98.28 \\

CAE~\citep{chen2024context}
& 0.31417 & 0.42417 & 76.66
& 0.17557 & 0.26151 & 89.42
& 40.08760 & 59.56808 & 97.74
& 87.15660 & 134.33110 & 98.59 \\

SimMIM~\citep{xie2022simmim}
& 0.28651 & 0.38220 & 81.05
& 0.15139 & 0.21604 & 92.78
& 39.90261 & \textbf{59.24226} & \textbf{97.76}
& 83.63551 & 138.52274 & 98.50 \\

Data2Vec-Base~\citep{baevski2022data2vec}
& 0.25115 & 0.37178 & 82.07
& 0.13043 & 0.21320 & 92.97
& 40.22185 & 60.93765 & 97.63
& 73.58009 & 124.37146 & 98.79 \\

Data2Vec-Large~\citep{baevski2022data2vec}
& 0.22278 & 0.33778 & 85.20
& 0.11900 & 0.19804 & 94.01
& 38.84260 & 59.28787 & 97.76
& 70.51602 & 121.79372 & 98.84 \\

MSN-Base~\citep{assran2022masked}
& 0.29429 & 0.42256 & 76.84
& 0.19804 & 0.31028 & 85.11
& 44.40505 & 64.83239 & 97.32
& 88.98394 & 144.31597 & 98.37 \\

MSN-Large~\citep{assran2022masked}
& 0.30508 & 0.42095 & 77.02
& 0.20727 & 0.30502 & 85.61
& 50.12944 & 70.63700 & 96.82
& 89.60778 & 142.97821 & 98.40 \\

\midrule

\textbf{SpectralTF-MAE-Base}
& 0.21281 & 0.32483 & 86.31
& 0.11008 & 0.19448 & 94.15
& 39.10889 & 61.79632 & 97.56
& \textbf{69.18898} & \textbf{121.00328} & \textbf{98.86} \\

\textbf{SpectralTF-MAE-Large}
& \textbf{0.19365} & \textbf{0.31588} & \textbf{87.06}
& \textbf{0.10532} & \textbf{0.19356} & \textbf{94.21}
& 42.79847 & 65.12049 & 97.30
& 77.02085 & 135.12647 & 98.57 \\

\bottomrule
\end{tabular}
}
\end{table}

Taken together, the Base and Large variants of SpectralTF-MAE achieve the best result on 9 of the 12 reported parameter-prediction metrics. SpectralTF-MAE-Large performs best on both electron-density quantities, obtaining the lowest MAE and RMSE and the highest $R^2$ for both $n_{e1}$ and $n_{e2}$. In contrast, SpectralTF-MAE-Base remains the strongest model for $T_e$, achieving the best MAE, RMSE, and $R^2$ in that target. For $T_i$, the best individual metrics are obtained by the competing baselines: ViT-Large gives the lowest MAE, while SimMIM gives the lowest RMSE and the highest $R^2$. These results show that the effect of scaling is target dependent: the Large configuration is particularly beneficial for electron-density prediction, whereas the Base configuration is better for $T_e$, and neither SpectralTF-MAE scale dominates the $T_i$ metrics.

For the compact ablation tables below, the mean parameter-prediction score is defined as
\begin{equation}
R^2_{\mathrm{Param}}
=
\frac{
R^2_{n_{e1}}+
R^2_{n_{e2}}+
R^2_{T_i}+
R^2_{T_e}
}{4}.
\end{equation}

\noindent $\bullet$ \textbf{Results on Spectral Peak Analysis.~}

Table~\ref{tab:peak_analysis} compares our task-specific peak-analysis baseline with SpectralTF-MAE using random and pretrained initialization.

\begin{table}[t]
\centering
\caption{Comparison on spectral peak analysis. Exact-$K$, F1, and weak-peak recall are reported in percentage. Reconstruction $R^2$ values are multiplied by 100.}
\label{tab:peak_analysis}
\footnotesize
\setlength{\tabcolsep}{2.0pt}
\renewcommand{\arraystretch}{1.12}
\begin{tabular*}{\textwidth}{@{}l@{\extracolsep{\fill}}cccccc@{}}
\toprule
Method &
Exact-$K$ $\uparrow$ &
F1 $\uparrow$ &
Weak Recall $\uparrow$ &
Center MAE $\downarrow$ &
Recon. $R^2$ $\uparrow$ &
NRMSE $\downarrow$ \\
\midrule
Task-specific Baseline
& 98.88
& 77.81
& 22.22
& 0.9631
& 98.54
& 0.02579 \\

SpectralTF-MAE-Base (Random)
& 98.92
& 80.67
& 19.05
& 0.9646
& \textbf{99.37}
& \textbf{0.01584} \\

SpectralTF-MAE-Base (Pretrained)
& \textbf{98.92}
& \textbf{83.64}
& \textbf{23.81}
& \textbf{0.8174}
& 99.22
& 0.01767 \\
\bottomrule
\end{tabular*}
\end{table}

Pretrained SpectralTF-MAE improves F1, weak-peak recall, and peak-center localization relative to both the task-specific baseline and the randomly initialized variant. The randomly initialized model, however, obtains slightly better waveform-level reconstruction metrics, including reconstruction $R^2$ and NRMSE. The effect of pretraining is therefore more evident in peak detection and localization than in uniform improvement across all reconstruction-oriented metrics.

\subsection{Ablation Study}

\begin{table}[!t]
\centering
\begin{minipage}[t]{0.49\textwidth}
\centering
\caption{Effect of pretrained initialization on representative downstream tasks.}
\label{tab:pretraining_ablation}
\scriptsize
\setlength{\tabcolsep}{1.2pt}
\renewcommand{\arraystretch}{1.03}
\resizebox{\linewidth}{!}{%
\begin{tabular}{lcccc}
\toprule
Model & Init. & \makecell{XCS$\rightarrow$CXRS\\Peak $R^2$} & \makecell{CXRS$\rightarrow$XCS\\Peak $R^2$} & \makecell{Param. Mean\\$R^2$} \\
\midrule
SpectralTF-MAE-Base & Random & 95.70 & 94.91 & 86.03 \\
\textbf{SpectralTF-MAE-Base} & \textbf{Pretrained} & \textbf{97.16} & \textbf{95.30} & \textbf{94.22} \\
\bottomrule
\end{tabular}}
\end{minipage}
\hfill
\begin{minipage}[t]{0.49\textwidth}
\centering
\caption{Component analysis of the peak-centered pretraining strategy.}
\label{tab:component_ablation}
\scriptsize
\setlength{\tabcolsep}{1.0pt}
\renewcommand{\arraystretch}{1.03}
\resizebox{\linewidth}{!}{%
\begin{tabular}{lcccc}
\toprule
Setting & \makecell{Peak-aware\\Masking} & \makecell{Peak-enhanced\\Loss} & \makecell{XCS$\rightarrow$CXRS\\$R^2$} & \makecell{Param. Mean\\$R^2$} \\
\midrule
Vanilla masked pretraining & \xmark & \xmark & 96.38 & 90.25 \\
+ Peak-enhanced loss & \xmark & \cmark & 96.71 & 91.69 \\
+ Peak-aware masking & \cmark & \xmark & 96.98 & 93.24 \\
\textbf{Full model} & \cmark & \cmark & \textbf{97.16} & \textbf{94.22} \\
\bottomrule
\end{tabular}}
\end{minipage}
\end{table}

\noindent $\bullet$ \textbf{Effect of Pretrained Initialization.~}

This experiment isolates the effect of self-supervised pretraining by comparing the same SpectralTF-MAE-Base architecture with random initialization and with the pretrained checkpoint. Spectral peak analysis is evaluated separately in Table~\ref{tab:peak_analysis}.

Pretrained initialization improves all three representative downstream scores relative to random initialization. The gain is largest for plasma-parameter prediction, with additional improvements in both cross-diagnostic generation directions. These results support the transferability of the learned spectral representation across distinct downstream tasks.

\noindent $\bullet$ \textbf{Component Analysis.~}

We further evaluate the two peak-centered components used during self-supervised pretraining: peak-aware masking and the peak-enhanced reconstruction loss. To limit experimental cost, XCS-to-CXRS generation and plasma-parameter prediction are used as representative downstream tasks.

Each peak-centered component improves the two representative downstream scores when added individually to vanilla masked pretraining. Combining peak-aware masking with the peak-enhanced loss yields the highest scores on both tasks, making the full configuration the best-performing combination among those evaluated.

\noindent $\bullet$ \textbf{Data Efficiency.~}

To limit experimental cost, data efficiency is evaluated on two representative downstream tasks: XCS-to-CXRS spectral generation and plasma-parameter prediction. Identical labeled-data subsets are used for the randomly initialized and pretrained variants, with $D_{10}\subset D_{25}\subset D_{50}\subset D_{100}$.

\begin{table}[H]
\centering
\caption{Data efficiency of spectral pretraining on representative downstream tasks.}
\label{tab:data_efficiency}
\setlength{\tabcolsep}{3.8pt}
\renewcommand{\arraystretch}{1.15}
\resizebox{0.82\columnwidth}{!}{
\begin{tabular}{c|cc|cc}
\toprule
\multirow{2}{*}{Labeled Data} &
\multicolumn{2}{c|}{XCS$\rightarrow$CXRS $R^2$} &
\multicolumn{2}{c}{Parameter Mean $R^2$} \\
\cline{2-5}
& Scratch & Pretrained
& Scratch & Pretrained \\
\midrule
10\%  & 84.88 & 95.94 & 58.72 & 69.60 \\
25\%  & 94.33 & 96.90 & 70.95 & 81.38 \\
50\%  & 95.56 & 97.03 & 79.59 & 89.23 \\
100\% & 95.70 & 97.16 & 86.03 & 94.22 \\
\bottomrule
\end{tabular}
}
\end{table}

Across all labeled-data fractions, the pretrained model outperforms the randomly initialized counterpart on both representative tasks. The gain is especially pronounced in the low-data regime for XCS-to-CXRS generation and remains clear for plasma-parameter prediction across all fractions, indicating improved label efficiency after self-supervised pretraining.

\subsection{Qualitative Results}
\begin{figure}[H]
    \vspace{-0.7em}
    \centering

    \begin{minipage}[t]{0.32\textwidth}
        \vspace{0pt}
        \centering
        \includegraphics[height=3.12cm,keepaspectratio]{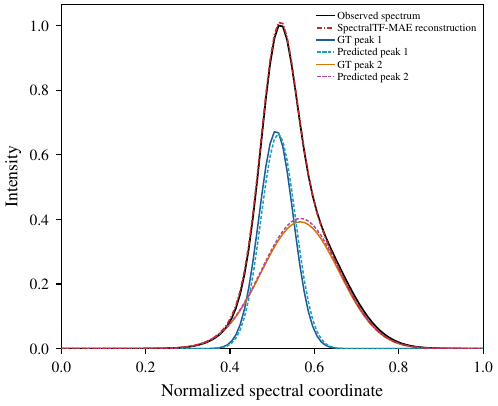}\\[-0.45mm]
        \parbox[t][1.00em][t]{\linewidth}{\centering\scriptsize (a) Spectral peak analysis}
    \end{minipage}
    \hspace{0.055\textwidth}
    \begin{minipage}[t]{0.50\textwidth}
        \vspace{0pt}
        \centering
        \includegraphics[height=3.12cm,keepaspectratio]{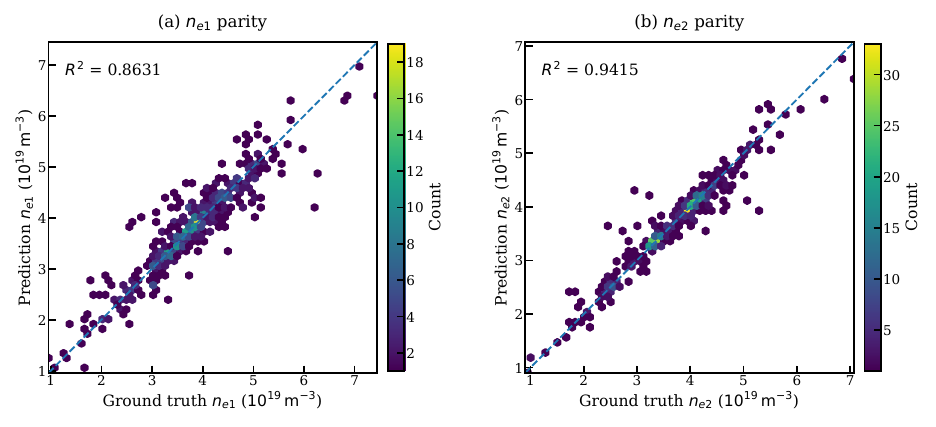}\\[-0.45mm]
        \parbox[t][1.00em][t]{\linewidth}{\centering\scriptsize (b) Electron-density prediction}
    \end{minipage}

    \vspace{-0.24em}

    \begin{minipage}[t]{0.46\textwidth}
        \vspace{0pt}
        \centering
        \makebox[\linewidth][c]{%
            \includegraphics[height=4.62cm,keepaspectratio]{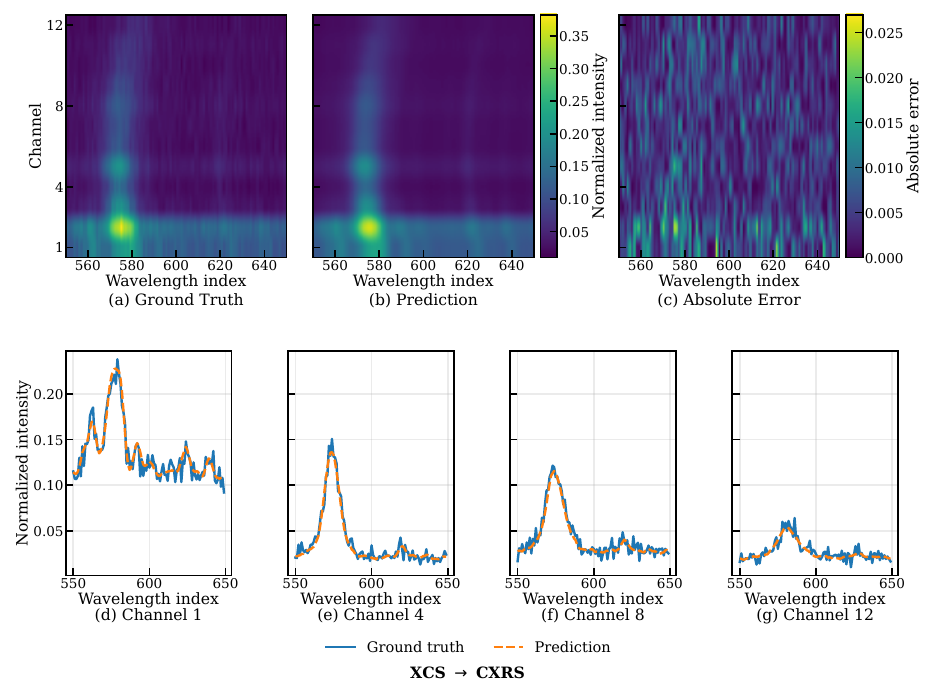}%
        }\\[-0.6mm]
        \parbox[t][1.35em][t]{\linewidth}{\centering\scriptsize (c) XCS-to-CXRS generation}
    \end{minipage}
    \hfill
    \begin{minipage}[t]{0.46\textwidth}
        \vspace{0pt}
        \centering
        \makebox[\linewidth][c]{%
            \includegraphics[height=4.62cm,keepaspectratio]{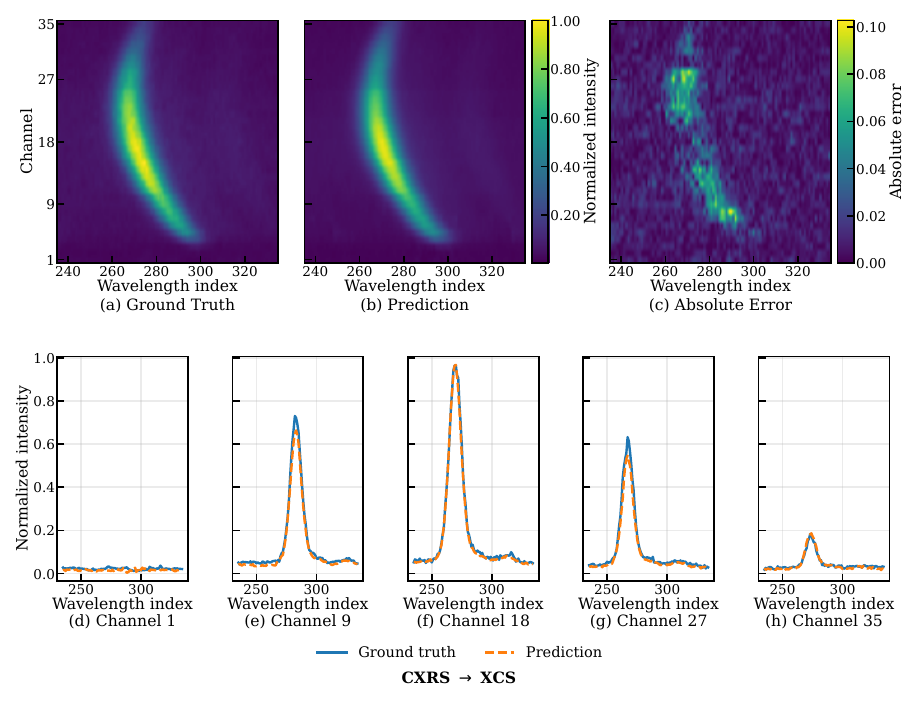}%
        }\\[-0.6mm]
        \parbox[t][1.35em][t]{\linewidth}{\centering\scriptsize (d) CXRS-to-XCS generation}
    \end{minipage}

    \caption{Representative qualitative results for the three downstream task families: (a) spectral peak decomposition, (b) electron-density prediction for $n_{e1}$ and $n_{e2}$, (c) XCS-to-CXRS spectral generation, and (d) CXRS-to-XCS spectral generation.}
    \label{fig:qualitative_main}
    \vspace{-1.05em}
\end{figure}

To complement the quantitative evaluation, Fig.~\ref{fig:qualitative_main} shows representative results from all three downstream task families. For spectral peak analysis, the reconstructed spectrum follows the observed spectrum and the predicted components remain consistent with the corresponding ground-truth peaks in position and overall line shape. For plasma-parameter prediction, the $n_{e1}$ and $n_{e2}$ parity plots show that most predictions lie close to the identity line over the observed density range. For cross-diagnostic generation, both XCS-to-CXRS and CXRS-to-XCS examples recover the dominant target-diagnostic response structures with localized residual errors. These complementary examples are consistent with the quantitative comparisons and illustrate transfer across spectral-component analysis, parameter inference, and bidirectional spectral generation. Additional qualitative results, including peak-center localization, temperature-profile errors, and the complete density analysis, are provided in Appendix~\ref{app:visualization}.

\vspace{-0.65em}
\section{Conclusion}
\vspace{-0.12em}

In this work, we presented SpectralTF-MAE, a self-supervised spectral foundation model for tokamak plasma diagnostics that learns transferable representations from paired CXRS and XCS measurements. By combining multi-diagnostic masked pretraining with peak-aware masking and a peak-enhanced reconstruction objective, the framework emphasizes diagnostically informative spectral structures while capturing shared correlations across the two modalities. The pretrained representation was evaluated on three complementary downstream tasks: bidirectional cross-diagnostic spectral generation, plasma-parameter prediction for $T_i$, $T_e$, and $n_e$, and spectral peak analysis. Experimental results on EAST data demonstrate strong transfer performance across these tasks, with particularly clear gains in cross-diagnostic generation and electron-density prediction, while pretraining also improves peak detection and localization. The ablation and data-efficiency studies further show that the proposed pretraining strategy provides consistent benefits over random initialization and remains effective when fewer labeled samples are available. Overall, these results support self-supervised spectral pretraining as a promising route toward reusable models for plasma spectroscopy and provide a foundation for extending spectral representation learning to broader diagnostic systems and fusion devices.

\section*{LLM Usage}
We used LLMs to improve the grammar, phrasing, and readability of this manuscript. We also used LLMs as a search tool to help us find related works, but we independently reviewed, interpreted, and summarized all identified works.

\bibliographystyle{plainnat}
\bibliography{reference}

\clearpage
\appendix

\section{Detailed Related Works}\label{app:related_work}

This section reviews prior work in three areas most closely related to our study: foundation models, artificial intelligence for nuclear fusion, and spectroscopy-based plasma diagnostics. Broader perspectives on machine learning for fusion and on diagnostic techniques for magnetic-confinement plasmas are available in existing reviews and community reports~\citep{humphreys2020advancing,weller2010diagnostics,mazon2025diagnostics}.

\subsection{Large Foundation Models}

Foundation models are typically pretrained on broad datasets to learn transferable representations that can be adapted to diverse downstream tasks~\citep{DBLP:journals/corr/abs-2108-07258}. The Transformer provides the self-attention architecture underlying many modern pretrained models~\citep{vaswani2017attention}, while BERT demonstrated the effectiveness of masked self-supervised pretraining for learning transferable language representations~\citep{devlin-etal-2019-bert}. This pretraining-and-adaptation paradigm was subsequently extended to vision through Vision Transformers~\citep{dosovitskiy2020image} and a range of masked or predictive self-supervised approaches, including MAE~\citep{He_2022_CVPR}, BEiT~\citep{bao2021beit}, SimMIM~\citep{xie2022simmim}, Context Autoencoder (CAE)~\citep{chen2024context}, Masked Siamese Networks (MSN)~\citep{assran2022masked}, data2vec~\citep{baevski2022data2vec}, masked spatiotemporal autoencoding~\citep{feichtenhofer2022masked}, and joint-embedding predictive learning~\citep{assran2023self}.

These ideas have also been extended to structured spectral and remote-sensing data. SpectralFormer introduced Transformer-based modeling for hyperspectral imagery~\citep{hong2021spectralformer}, while SatMAE extended masked pretraining to temporal and multispectral satellite observations~\citep{cong2022satmae}. Subsequent studies investigated masked vision Transformers for hyperspectral classification~\citep{scheibenreif2023masked}, spatial--spectral masked autoencoding~\citep{lin2023ss}, factorized hyperspectral Transformers with self-supervised pretraining~\citep{mohamed2023factoformer}, and spectral masked autoencoding for few-shot hyperspectral learning~\citep{feng2023spectral}. More recently, SpectralGPT~\citep{10490262} and HyperSIGMA~\citep{wang2025hypersigma} developed foundation-model approaches for spectral remote sensing. Beyond remote sensing, AstroCLIP explored cross-modal foundation modeling for astronomical observations~\citep{parker2024astroclip}, while Leung and Bovy developed a self-supervised Transformer for stellar spectra and physical-parameter inference~\citep{leung2024towards}. Collectively, these studies demonstrate the potential of reusable representations for high-dimensional spectral measurements; however, foundation modeling for plasma spectroscopy remains comparatively underexplored.

\subsection{AI for Nuclear Fusion}

Machine learning has been applied extensively to magnetic-confinement fusion for disruption prediction, diagnostic inference, and plasma control~\citep{humphreys2020advancing}. Representative disruption studies include machine-learning approaches on DIII-D and Alcator C-Mod~\citep{rea2018disruption}, multi-machine warning systems spanning Alcator C-Mod, DIII-D, and EAST~\citep{montes2019machine}, real-time prediction on DIII-D~\citep{rea2019real}, and deep convolutional models using high-temporal-resolution diagnostic signals~\citep{churchill2020deep}. Deep learning has also been applied to plasma tomography and disruption prediction from bolometer measurements~\citep{ferreira2019deep}, while cross-device approaches have demonstrated the potential for generalization in disruptive-instability prediction~\citep{kates2019predicting}.

Beyond prediction, learning-based methods have been extended to plasma control and operation optimization. Deep reinforcement learning has been demonstrated for magnetic control of tokamak plasmas~\citep{degrave2022magnetic} and for avoidance of tearing instabilities~\citep{seo2024avoiding}; recent work has also integrated machine learning into high-performance tokamak operation strategies~\citep{kim2024highest}. For diagnostic inference, deep-learning models have been developed to estimate quantities such as upstream separatrix density at JET~\citep{kit2023developing}. These studies illustrate the breadth of AI applications in fusion experiments, although most models remain tailored to individual tasks or diagnostic channels.

More recently, reusable pretrained models have attracted growing attention in fusion research. Churchill~\citeyearpar{churchill2025ai} discussed the potential of foundation models for exploiting unlabeled fusion data, while FusionMAE~\citep{yang2026fusionmae} introduced masked self-supervised pretraining over multiple HL-3 diagnostic and operational signals and transferred the learned representation to several downstream tasks. Recent fusion foundation-model studies have primarily emphasized heterogeneous diagnostic time series and global plasma-state representations. In contrast, our work focuses on high-dimensional XCS and CXRS measurements and investigates whether a shared spectral representation can transfer across plasma-parameter inference and cross-diagnostic spectral generation.

\subsection{Spectroscopy-Based Plasma Diagnostics}

Spectroscopic diagnostics provide key measurements of electron and ion temperatures, density, impurity states, and plasma rotation in magnetic-confinement experiments~\citep{weller2010diagnostics,mazon2025diagnostics}. Charge-exchange recombination spectroscopy (CXRS) infers ion temperature and rotation from Doppler broadening and spectral-line shifts~\citep{fonck1984determination,isler1994overview}. CXRS systems, together with calibration and high-speed acquisition schemes, have been developed on EAST~\citep{li2014development,ye2015toroidal,zhang2015calibration,li2019concept}, and related high-resolution spectroscopic diagnostics have been implemented on devices such as COMPASS~\citep{weinzettl2015high}.

X-ray crystal spectroscopy (XCS/XICS) provides complementary high-resolution measurements of impurity X-ray emission. Spatially resolved crystal spectrometers have been developed to measure ion and electron temperatures and plasma rotation~\citep{bitter2004spatially,ince2008spatially,hill2008spatially}. Related systems have been implemented on KSTAR~\citep{lee2004research}, EAST~\citep{shi2010imaging,lyu2014upgrades}, and HL-3~\citep{zhang2024x}, and X-ray imaging crystal spectroscopy has also been applied to plasma-transport studies~\citep{reinke2012x}.

Conventional spectroscopic analysis typically relies on spectral fitting, line-ratio analysis, atomic models, or inversion procedures. Machine learning has therefore been investigated as a data-driven alternative for rapid spectral inference. Chai et al.~\citeyearpar{chai2019fast} used neural networks to estimate ion temperature from EAST CXRS measurements, while Samuell et al.~\citeyearpar{samuell2021measuring} inferred electron temperature and plasma detachment from EUV/VUV spectra. Lin et al. developed neural-network models for predicting ion temperature and plasma rotation from EAST XCS measurements~\citep{lin2024prediction} and subsequently investigated electron-temperature inference from XCS spectra~\citep{lin2025development}. Neural-network-based analysis of charge-exchange spectra has also been demonstrated on KSTAR~\citep{lee2026neural}. Outside fusion, deep neural networks have been applied to automatic peak picking and spectral deconvolution in NMR and high-resolution LC--MS data~\citep{li2021deep,melnikov2020deep}, further illustrating the potential of learning-based spectral-structure analysis.

Despite these advances, existing AI-based plasma-spectroscopy methods are predominantly task-specific and fully supervised, with separate models trained for individual diagnostic quantities. Consequently, spectral structures shared across diagnostics and downstream tasks are not explicitly exploited. In contrast, this work learns a reusable representation from paired XCS and CXRS measurements through masked self-supervised pretraining and transfers the pretrained model to multiple downstream tasks, including electron-density prediction, ion- and electron-temperature prediction, and cross-diagnostic spectral generation.

\section{Pretraining Details}\label{app:pretraining_details}

The pretraining stage learns a shared spectral representation from paired CXRS and XCS measurements without plasma-parameter labels. As shown in Fig.~\ref{fig:framework}, both spectra are normalized and converted into modality-specific tokens. Peak-aware masking defines the reconstruction targets, while the remaining visible tokens are jointly encoded by a shared Transformer. Modality-specific decoders then reconstruct the two diagnostics from the shared latent representation, and a peak-enhanced objective places additional emphasis on the characteristic spectral regions.

\subsection{Multi-diagnostic Spectral Tokenization}

Let $\mathbf{X}^{c}\in\mathbb{R}^{H_c\times W_c}$ and $\mathbf{X}^{x}\in\mathbb{R}^{H_x\times W_x}$ denote paired CXRS and XCS measurements, respectively, with the two dimensions corresponding to the sampling-position and wavelength axes. Each spectrum is independently normalized using sample-wise min--max normalization:
\begin{equation}
    \widetilde{\mathbf{X}}^{m}
    =
    \frac{\mathbf{X}^{m}-\min_{h,w}X^{m}_{h,w}}
    {\max_{h,w}X^{m}_{h,w}-\min_{h,w}X^{m}_{h,w}+\epsilon},
    \qquad m\in\{c,x\}.
\end{equation}

Unlike natural images, the spectral tensors are anisotropic: their two axes represent distinct physical dimensions and have substantially different sampling resolutions. We therefore use modality-specific rectangular patches elongated along the wavelength axis, following the general patch-based Transformer formulation used in ViT and spectral Transformer models~\citep{dosovitskiy2020image,hong2021spectralformer} while adapting the patch geometry to the sampling characteristics of CXRS and XCS.

For each modality, a modality-specific input adapter projects spectral patches into a common latent space, and a two-dimensional positional embedding encodes their sampling-position and wavelength locations. The $i$-th token is written as
\begin{equation}
    \mathbf{t}^{m}_{i}
    =
    \mathcal{E}_{m}\left(\widetilde{\mathbf{X}}^{m}_{i}\right)
    +\mathbf{p}^{m}_{i},
\end{equation}
where $\mathcal{E}_{m}$ denotes the modality-specific patch projection and $\mathbf{p}^{m}_{i}$ is the corresponding positional embedding. Although CXRS and XCS use different patch geometries, both are mapped into the same latent space for joint modeling.

\subsection{Peak-aware Spectral Masking}

Uniform random masking assigns the same selection probability to all spectral patches, whereas the dominant diagnostic responses in the considered plasma spectra are concentrated around characteristic emission peaks. We therefore introduce peak-aware masking, which preferentially includes high-response patches among the reconstruction targets while retaining part of the peak-related information as visible context.

For each patch, its maximum normalized intensity is used as the peak score:
\begin{equation}
    s_i^{m}
    =
    \max_{(h,w)\in\mathcal{P}^{m}_{i}}
    \widetilde{X}^{m}_{h,w},
\end{equation}
where $\mathcal{P}^{m}_{i}$ denotes the $i$-th patch of modality $m$. For each paired sample, the CXRS and XCS scores are concatenated and used to compute a sample-specific quantile threshold:
\begin{equation}
    \tau=Q_q\left([\mathbf{s}^{c};\mathbf{s}^{x}]\right).
\end{equation}
Patches with scores above $\tau$ are regarded as peak candidates.

During masking, peak candidates are preferentially selected as reconstruction targets, while the fraction of masked peak candidates is constrained so that part of the characteristic spectral information remains visible to the encoder. The remaining masking positions are sampled from non-peak tokens. This design increases the reconstruction emphasis on informative spectral structures without removing all peak-related context.

\subsection{Joint Spectral Representation and Reconstruction}

After masking, the visible CXRS and XCS tokens are concatenated and processed by a shared Transformer encoder. A learnable global token is included to aggregate information across the paired diagnostics:
\begin{equation}
    \mathbf{Z}
    =
    \mathcal{F}_{\mathrm{enc}}
    \left(
    [\mathcal{V}^{c};\mathcal{V}^{x};\mathbf{g}]
    \right),
\end{equation}
where $\mathcal{V}^{c}$ and $\mathcal{V}^{x}$ denote the visible token sets and $\mathcal{F}_{\mathrm{enc}}$ is the shared encoder. Because tokens from both diagnostics interact within the same self-attention layers, the encoder can jointly model intra-diagnostic spectral dependencies and cross-diagnostic correlations.

Separate output adapters are used to reconstruct CXRS and XCS:
\begin{equation}
    \widehat{\mathbf{X}}^{c}
    =\mathcal{D}_{c}(\mathbf{Z}),
    \qquad
    \widehat{\mathbf{X}}^{x}
    =\mathcal{D}_{x}(\mathbf{Z}).
\end{equation}
For each diagnostic, mask tokens and positional information are restored at their original patch locations. Modality-specific reconstruction queries then attend to the shared encoder representation through cross-attention:
\begin{equation}
    \mathbf{Q}^{m}_{\mathrm{att}}
    =
    \operatorname{CrossAttn}
    \left(\mathbf{Q}^{m},\mathbf{Z}\right),
    \qquad m\in\{c,x\}.
\end{equation}
The resulting query features are further refined by the decoder and projected back to spectral patches. The two diagnostics therefore share a common latent representation while retaining modality-specific reconstruction pathways.

\subsection{Peak-enhanced Reconstruction Objective}

For each diagnostic, the basic reconstruction objective is the mean squared error computed over the masked spectral region. Let $\mathcal{M}^{m}$ denote the two-dimensional region corresponding to the masked patches:
\begin{equation}
    \mathcal{L}_{\mathrm{rec}}^{m}
    =
    \frac{1}{|\mathcal{M}^{m}|}
    \sum_{(h,w)\in\mathcal{M}^{m}}
    \left(
    \widehat{X}^{m}_{h,w}
    -
    \widetilde{X}^{m}_{h,w}
    \right)^2.
\end{equation}

Because the characteristic emission response occupies only a subset of the wavelength axis, we additionally define a characteristic wavelength-index interval $I_m$ for each diagnostic. The corresponding two-dimensional peak region is
\begin{equation}
    \Omega_m
    =
    \left\{
    (h,w)\mid
    w\in I_m
    \right\}.
\end{equation}
The peak-enhanced loss is computed over the intersection of the characteristic region and the masked reconstruction region:
\begin{equation}
    \mathcal{L}_{\mathrm{peak}}^{m}
    =
    \frac{1}{|\Omega_m\cap\mathcal{M}^{m}|}
    \sum_{(h,w)\in\Omega_m\cap\mathcal{M}^{m}}
    \left(
    \widehat{X}^{m}_{h,w}
    -
    \widetilde{X}^{m}_{h,w}
    \right)^2.
\end{equation}
If no masked element falls within the predefined characteristic region for a given sample and modality, the corresponding peak-loss term is set to zero.

The overall pretraining objective combines the masked reconstruction loss with the peak-enhanced term:
\begin{equation}
    \mathcal{L}_{\mathrm{pre}}
    =
    \sum_{m\in\{c,x\}}
    \left(
    \mathcal{L}_{\mathrm{rec}}^{m}
    +
    \lambda_{\mathrm{peak}}
    \mathcal{L}_{\mathrm{peak}}^{m}
    \right).
\end{equation}
Peak-aware masking determines which spectral structures are preferentially selected for reconstruction, whereas the peak-enhanced loss assigns additional optimization weight to predefined characteristic regions. Together, these components constitute the peak-centered self-supervised objective used for multi-diagnostic spectral representation learning.

\section{Downstream Task Details}\label{app:downstream_details}

After self-supervised pretraining, the learned representation is adapted to downstream diagnostic tasks through task-specific fine-tuning. We consider three task families: plasma-parameter prediction from XCS measurements, including temperature-profile and electron-density inference; cross-diagnostic spectral generation between XCS and CXRS; and spectral peak analysis. Together, these tasks evaluate the pretrained representation in terms of physical-parameter inference, cross-diagnostic correspondence, and spectral-component analysis.

\noindent $\bullet$ \textbf{Plasma-Parameter Prediction.~}

Spectroscopic diagnostics provide an important route for obtaining plasma kinetic parameters. In particular, ion temperature ($T_i$) and electron temperature ($T_e$) are fundamental for characterizing plasma confinement, turbulent transport, and stability, and established spectroscopic analysis can recover temperature information from spectral-line shapes and intensity responses~\citep{bitter2004spatially,ince2008spatially,hill2008spatially}. However, conventional spectral fitting and inversion procedures can be computationally expensive for large-scale or rapid analysis. Recent neural-network approaches have therefore learned direct mappings from EAST spectra to $T_i$ and $T_e$ profiles~\citep{lin2024prediction,lin2025development}, demonstrating the feasibility of fast data-driven temperature inference. These approaches are typically developed and trained separately for individual targets, however, and do not provide a reusable pretrained spectral representation.

Electron-density prediction constitutes a complementary and more exploratory task. Unlike temperature quantities with established spectroscopic signatures, electron density is not determined from raw XCS detector counts through a comparably direct mapping; its spectral dependence is entangled with line intensities and other plasma and atomic conditions. This makes conventional direct inversion from XCS measurements to $n_e$ difficult. We therefore investigate whether a data-driven model can learn these latent correlations directly from experimental measurements. Accordingly, our plasma-parameter benchmark contains two complementary settings: established temperature-profile prediction for $T_i$ and $T_e$, and exploratory electron-density prediction for $n_e$.

For temperature prediction, both $T_i$ and $T_e$ are formulated as radial-profile regression targets, with each profile represented at 32 radial positions. For electron density, two measurements are available: $n_{e1}$ corresponds to the solid-source measurement, whereas $n_{e2}$ is obtained from POINT. Their consistency is used for data screening, and samples showing substantial disagreement between the two measurements are excluded as invalid. After screening, a total of 5,330 valid samples are retained for the downstream plasma-parameter prediction task.

For each valid sample, the complete XCS spectrum is processed by the pretrained spectral encoder, while the reconstruction decoders used during pretraining are removed. The global representation and local XCS tokens are pooled and concatenated into a fused feature vector $\mathbf{f}_{x}$, which is passed to independent task-specific regression heads:
\begin{equation}
\widehat{\mathbf{y}}_{k}
=
\mathcal{H}_{k}(\mathbf{f}_{x}),
\qquad
k\in\{n_{e1},n_{e2},T_i,T_e\},
\end{equation}
where $\mathcal{H}_{k}$ denotes the task-specific regression head for either electron-density measurement, ion-temperature profiles, or electron-temperature profiles.

This task tests whether the self-supervised spectral representation encodes plasma-state information that is useful for quantitative parameter inference. Under the same downstream training setting, pretrained initialization improves the mean $R^2$ for plasma-parameter prediction from 86.03 to 94.22 (values multiplied by 100), corresponding to an absolute improvement of 8.19 percentage points over random initialization.

\noindent $\bullet$ \textbf{Cross-Diagnostic Spectral Generation.~}

We next evaluate whether the pretrained representation captures transferable correlations between spectroscopic diagnostics through cross-diagnostic spectral generation. Given a complete source spectrum, all source tokens are passed to the pretrained encoder without masking, and the target-diagnostic decoder reconstructs the complete target spectrum:
\begin{equation}
\widehat{\mathbf{X}}^{t}
=
\mathcal{D}_{t}
\left(
\mathcal{F}_{\mathrm{enc}}(\mathbf{X}^{s})
\right),
\qquad s\neq t,
\end{equation}
where $s,t\in\{x,c\}$ denote the source and target diagnostics, respectively.

Although the decoder produces the complete target spectral tensor, optimization is restricted to the predefined characteristic peak region of the target diagnostic. Let $\Omega_t$ denote the characteristic region defined during pretraining. The downstream spectral-generation loss is
\begin{equation}
\mathcal{L}_{\mathrm{gen}}^{s\rightarrow t}
=
\frac{1}{|\Omega_t|}
\sum_{(h,w)\in\Omega_t}
\left(
\widehat{X}^{t}_{h,w}
-
\widetilde{X}^{t}_{h,w}
\right)^2.
\end{equation}
This formulation is consistent with the peak-centered pretraining strategy and concentrates downstream supervision on the spectral structures of primary diagnostic interest.

Both XCS-to-CXRS and CXRS-to-XCS directions are fine-tuned independently, providing complementary tests of whether the pretrained representation captures cross-diagnostic correspondence beyond modality-specific self-reconstruction.

\noindent $\bullet$ \textbf{Spectral Peak Analysis.~}

We further evaluate representation transfer on a spectral peak-analysis task, where the downstream model decomposes an input spectrum into multiple components and predicts the center, amplitude, full width at half maximum (FWHM), and existence probability of each peak. To isolate the contribution of pretraining, we retain the original task-specific peak-analysis pipeline and optimization objectives, replacing only its Transformer encoder with the pretrained SpectralTF encoder and fine-tuning it jointly with the downstream model. This controlled transfer setting tests whether representations learned from CXRS--XCS pretraining generalize to a structurally different spectral-analysis task.

\section{Dataset and Evaluation Details}\label{app:data_metrics}
\subsection{Datasets and Preprocessing}

The spectral dataset used in this study was constructed from CXRS and XCS measurements acquired during EAST experiments from 2021 to 2023. Only measurements for which valid CXRS and XCS data were simultaneously available were selected and paired according to the corresponding discharge and acquisition time. After data synchronization and quality control, a total of 15303 valid paired samples were retained for subsequent model development.

These paired measurements cover a broad range of EAST plasma conditions and provide the experimental basis for spectral pretraining and downstream evaluation. The complete CXRS input is represented as a $1024\times12$ spectral matrix. Because only 8 CXRS channels were available for part of the dataset owing to experimental-configuration changes, these incomplete measurements are reconstructed using a separately trained channel-completion model and then combined with the complete-channel samples. XCS measurements are spatially integrated to a compact $35\times487$ representation.

After quality control, data augmentation is applied only to complete-channel samples in the training set. Paired CXRS and XCS spectra are shifted jointly along the wavelength direction to preserve cross-diagnostic correspondence, and Poisson noise is added to emulate measurement fluctuations. Validation and test samples, as well as channel-completed samples, are not augmented, preventing augmentation-induced information leakage across data splits. The resulting paired CXRS--XCS dataset is used for self-supervised spectral pretraining and downstream evaluation.

\subsection{Evaluation Metrics}

We evaluate the regression and spectral-generation tasks using three standard metrics: mean absolute error (MAE), root mean squared error (RMSE), and the coefficient of determination ($R^2$). These metrics characterize average prediction error, sensitivity to larger deviations, and agreement between predicted and ground-truth values, respectively.

Given $N$ target values, with $y_i$ and $\widehat{y}_i$ denoting the ground truth and prediction, respectively, the MAE is defined as
\begin{equation}
\mathrm{MAE}
=
\frac{1}{N}
\sum_{i=1}^{N}
\left|
y_i-\widehat{y}_i
\right|.
\end{equation}

The RMSE is defined as
\begin{equation}
\mathrm{RMSE}
=
\sqrt{
\frac{1}{N}
\sum_{i=1}^{N}
\left(
y_i-\widehat{y}_i
\right)^2
}.
\end{equation}
Because the error is squared before averaging, RMSE places greater weight on large deviations than MAE.

The coefficient of determination is calculated as
\begin{equation}
R^2
=
1-
\frac{
\sum_{i=1}^{N}
\left(
y_i-\widehat{y}_i
\right)^2
}{
\sum_{i=1}^{N}
\left(
y_i-\overline{y}
\right)^2
},
\end{equation}
where $\overline{y}$ is the mean of the ground-truth values. Higher $R^2$ and lower MAE/RMSE indicate better predictive agreement. For readability, all $R^2$ values reported in the following experiments are multiplied by 100.

For plasma-parameter prediction, model outputs are transformed from the standardized target space back to their original physical scales before evaluation. The $T_i$ and $T_e$ profiles each contain 32 radial positions; for each profile quantity, predictions from all test samples and radial positions are aggregated before the metrics are computed.

For cross-diagnostic spectral generation, metrics are computed in the normalized spectral space and restricted to the predefined characteristic region of the target diagnostic. Specifically, the CXRS characteristic region is $\Omega_c=[550,650)$ and the XCS characteristic region is $\Omega_x=[235,335)$. Predicted and ground-truth values from all test samples, sampling positions, and wavelengths within the corresponding region are aggregated to compute MAE, RMSE, and $R^2$. This protocol focuses the quantitative assessment on the spectral structures of primary diagnostic interest rather than the non-characteristic background.

For spectral peak analysis, we report Precision, Recall, F1 score, Exact-$K$ accuracy, center MAE, amplitude normalized MAE (NMAE), FWHM relative error, and the reconstruction $R^2$ between the reconstructed spectrum and the ground-truth spectrum. Among these metrics, F1 and Exact-$K$ quantify peak-detection performance, whereas the center, amplitude, and FWHM errors measure the accuracy of the estimated spectral-peak parameters.

\section{Additional Ablation Studies}\label{app:additional_ablation}
\noindent $\bullet$ \textbf{Effect of Spectral Patch Size.~}

We compare representative rectangular patch configurations while holding the remaining pretraining settings fixed. For each configuration, the visible-token budget is adjusted to maintain an approximately 80\% global masking ratio. XCS patch sizes are first compared with the CXRS patch fixed; the CXRS patch size is then evaluated using the selected XCS configuration.

\begin{table}[t]
\centering
\caption{Effect of XCS spectral patch size on representative downstream tasks.
All $R^2$ values are multiplied by 100.}
\label{tab:xcs_patch}
\setlength{\tabcolsep}{4.5pt}
\renewcommand{\arraystretch}{1.15}
\resizebox{\columnwidth}{!}{
\begin{tabular}{lccc}
\toprule
XCS Patch &
XCS$\rightarrow$CXRS $R^2$ $\uparrow$ &
CXRS$\rightarrow$XCS $R^2$ $\uparrow$ &
Parameter Mean $R^2$ $\uparrow$ \\
\midrule
$5\times10$  & \textbf{97.20} & 94.92 & 92.88 \\
$7\times8$   & 97.11 & 95.05 & 92.53 \\
$7\times10$  & 96.55 & 94.82 & 93.24 \\
$5\times14$  & 97.16 & \textbf{95.30} & \textbf{94.22} \\
$7\times14$  & 96.89 & 95.15 & 92.99 \\
\bottomrule
\end{tabular}
}
\end{table}

\begin{table}[t]
\centering
\caption{Effect of CXRS patch size on downstream transfer. The XCS patch size is fixed to the configuration selected from the XCS patch study.}
\label{tab:cxrs_patch}
\setlength{\tabcolsep}{4.5pt}
\renewcommand{\arraystretch}{1.15}
\resizebox{\columnwidth}{!}{
\begin{tabular}{lccc}
\toprule
CXRS Patch &
XCS$\rightarrow$CXRS $R^2$ $\uparrow$ &
CXRS$\rightarrow$XCS $R^2$ $\uparrow$ &
Parameter Mean $R^2$ $\uparrow$ \\
\midrule
$6\times8$  & 96.80 & 95.21 & 93.27 \\
$3\times16$ & \textbf{97.16} & \textbf{95.30} & \textbf{94.22} \\
\bottomrule
\end{tabular}
}
\end{table}

For XCS, the $5\times10$ patch gives a marginally higher XCS-to-CXRS $R^2$ than $5\times14$ (97.20 versus 97.16), whereas $5\times14$ achieves the best CXRS-to-XCS and parameter-prediction scores. We therefore select $5\times14$ as the default XCS configuration because it provides the best overall transfer performance across the evaluated tasks rather than the best score on every individual task. With the XCS patch fixed at $5\times14$, the $3\times16$ CXRS patch outperforms $6\times8$ on all three reported downstream metrics.

\noindent $\bullet$ \textbf{Sensitivity to Global Masking Ratio.~}

\begin{table}[t]
\centering
\caption{Sensitivity to the global masking ratio during spectral pretraining.}
\label{tab:mask_ratio}
\footnotesize
\renewcommand{\arraystretch}{1.12}
\begin{tabular*}{\columnwidth}{@{}c@{\extracolsep{\fill}}cc@{}}
\toprule
Mask Ratio &
XCS$\rightarrow$CXRS $R^2$ $\uparrow$ &
Parameter Mean $R^2$ $\uparrow$ \\
\midrule
60\% & 97.04 & 91.72 \\
70\% & 97.18 & 92.59 \\
80\% & 97.16 & 94.22 \\
90\% & 97.12 & 92.28 \\
\bottomrule
\end{tabular*}
\end{table}

A 70\% masking ratio gives a marginally higher XCS-to-CXRS $R^2$ than 80\% (97.18 versus 97.16), whereas 80\% yields the highest parameter-prediction score. We therefore adopt 80\% as the default because it provides the best overall trade-off across the evaluated downstream tasks rather than being uniformly optimal for every task.

\noindent $\bullet$ \textbf{Effect of Pretraining Data Augmentation.~}

Data augmentation is applied only during self-supervised pretraining; the fine-tuning, validation, and test data are unchanged. Both cross-diagnostic generation directions are used to assess its effect on downstream transfer.

\begin{table}[t]
\centering
\caption{Effect of data augmentation during self-supervised pretraining on cross-diagnostic spectral generation. MAE and RMSE are calculated in the normalized target characteristic region, and $R^2$ values are multiplied by 100.}
\label{tab:spectral_aug}
\setlength{\tabcolsep}{5.0pt}
\renewcommand{\arraystretch}{1.15}
\resizebox{\linewidth}{!}{
\begin{tabular}{c|ccc|ccc}
\toprule
\multirow{2}{*}{Pretraining Aug.} &
\multicolumn{3}{c|}{XCS $\rightarrow$ CXRS} &
\multicolumn{3}{c}{CXRS $\rightarrow$ XCS} \\
\cline{2-7}
& MAE $\downarrow$
& RMSE $\downarrow$
& $R^2$ $\uparrow$
& MAE $\downarrow$
& RMSE $\downarrow$
& $R^2$ $\uparrow$ \\
\midrule
No
& 0.00809
& 0.01615
& 95.86
& 0.02080
& 0.03417
& 94.86 \\

\textbf{Yes}
& \textbf{0.00632}
& \textbf{0.01339}
& \textbf{97.16} 
& \textbf{0.01796}
& \textbf{0.03269}
& \textbf{95.30} \\
\bottomrule
\end{tabular}
}
\end{table}

Pretraining augmentation improves MAE, RMSE, and $R^2$ in both cross-diagnostic generation directions. The augmentation strategy is therefore retained in the default pretraining configuration.

\section{Additional Qualitative Results}\label{app:visualization}

To complement the quantitative evaluation, we visualize representative results for cross-diagnostic spectral generation, spectral peak analysis, and plasma-parameter prediction.

\noindent $\bullet$ \textbf{Cross-Diagnostic Spectral Generation.~}

\begin{figure}[!t]
    \centering
    \subfigure[XCS-to-CXRS generation.]{
        \includegraphics[width=0.47\textwidth]{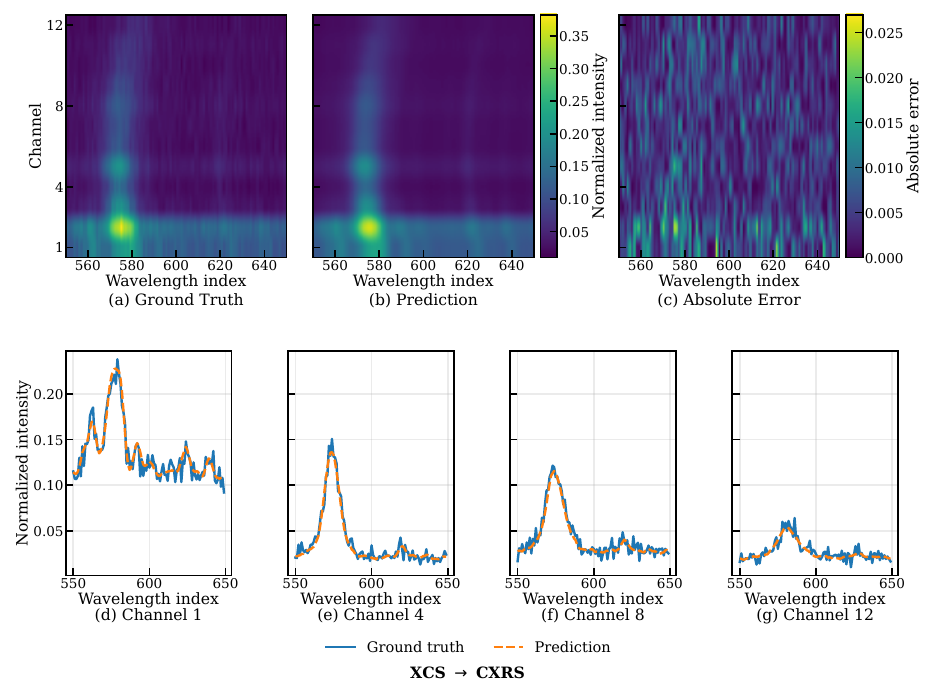}
    }
    \hfill
    \subfigure[CXRS-to-XCS generation.]{
        \includegraphics[width=0.47\textwidth]{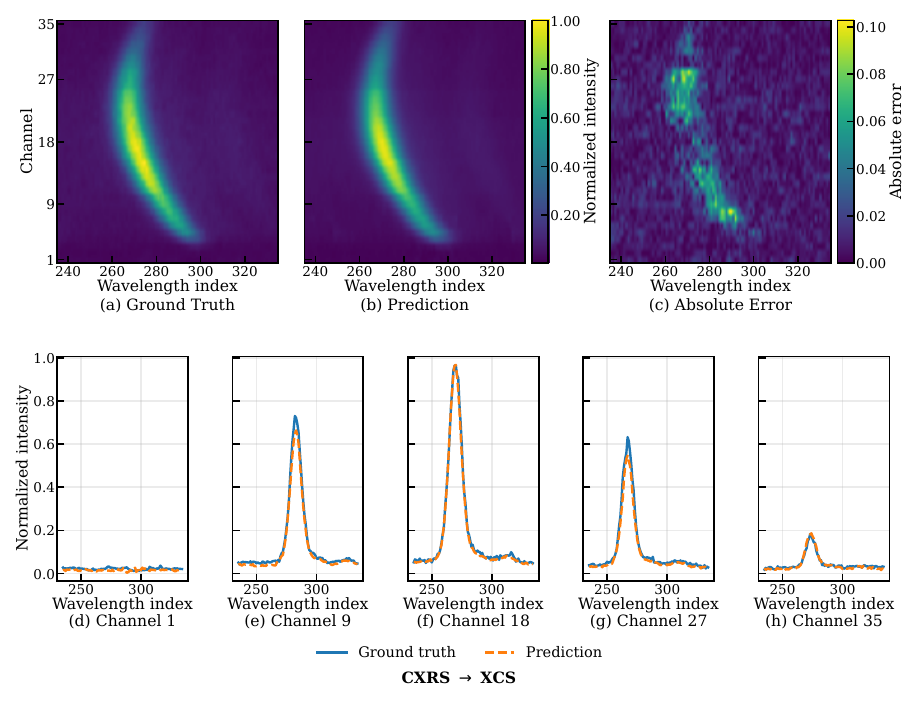}
    }
    \caption{Visualization of bidirectional spectral generation: (a) XCS-to-CXRS generation and (b) CXRS-to-XCS generation. Each subfigure compares the ground-truth target spectrum, the generated spectrum, the absolute-error map, and representative spectral profiles.}
    \label{fig:bidirectional_generation}
\end{figure}

Fig.~\ref{fig:bidirectional_generation} presents representative generation results for both cross-diagnostic directions. Fig.~\ref{fig:bidirectional_generation}~(a) shows XCS-to-CXRS generation, while Fig.~\ref{fig:bidirectional_generation}~(b) shows the reverse CXRS-to-XCS generation. In both subfigures, the visualization compares the ground-truth target spectrum, the generated spectrum, and the corresponding absolute-error map, together with one-dimensional profiles at representative sampling positions.

As shown in Fig.~\ref{fig:bidirectional_generation}~(a), the generated CXRS spectrum reproduces the dominant structures of the ground-truth spectrum, including the locations and overall shapes of the characteristic emission responses across sampling positions. The representative profiles also show close agreement in peak intensity and local line shape, while the remaining discrepancies are mainly localized around several strong responses.

Fig.~\ref{fig:bidirectional_generation}~(b) presents the reverse CXRS-to-XCS direction. The generated XCS spectrum recovers the characteristic curved emission structure across the sampling dimension, and the representative profiles broadly agree in peak position, intensity, and local shape. Taken together, the two subfigures are consistent with the quantitative results in Table~\ref{tab:bidirectional_generation} and support transferable correspondence between the two diagnostics.

\noindent $\bullet$ \textbf{Spectral Peak Analysis.~}

\begin{figure}[!t]
    \centering
    \begin{minipage}[t]{0.48\textwidth}
        \centering
        \parbox[c][6.0cm][c]{\linewidth}{\centering\includegraphics[width=\linewidth,height=5.9cm,keepaspectratio]{figures/peak_decomposition.pdf}}
        \caption{Visualization of spectral peak decomposition for a representative sample. The observed spectrum and SpectralTF-MAE reconstruction are shown together with the ground-truth peak components and their corresponding predicted components.}
        \label{fig:peak_decomposition}
    \end{minipage}
    \hfill
    \begin{minipage}[t]{0.48\textwidth}
        \centering
        \parbox[c][6.0cm][c]{\linewidth}{\centering\includegraphics[width=\linewidth,height=5.9cm,keepaspectratio]{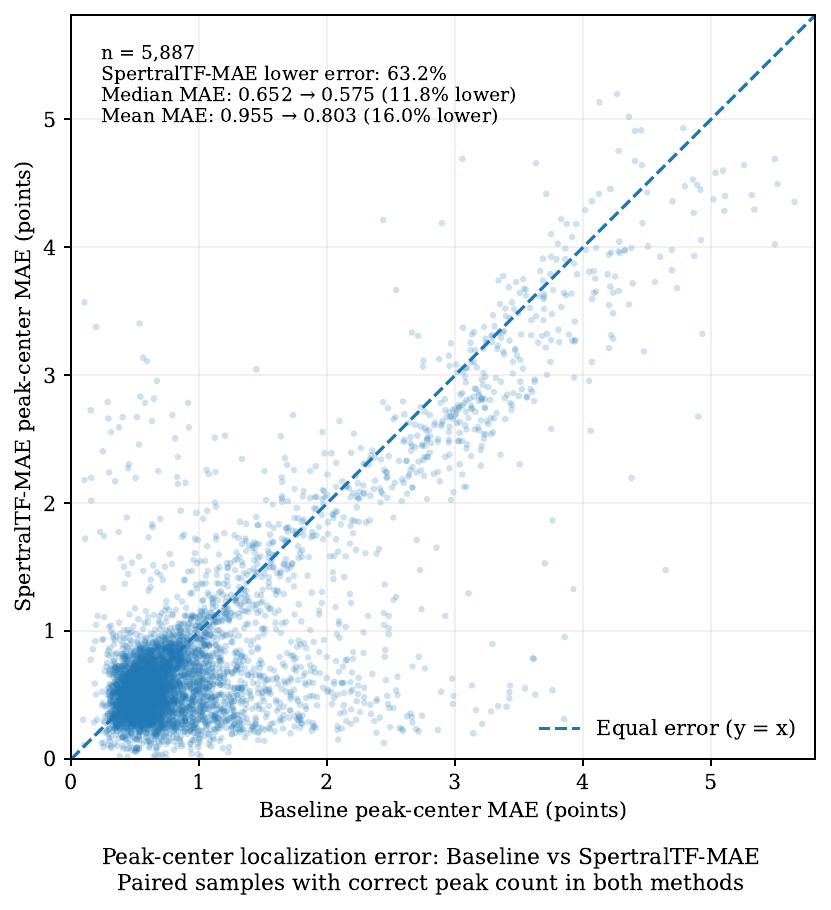}}
        \caption{Peak-center localization error comparison between the baseline method and SpectralTF-MAE. Each point denotes a paired sample with the correct peak count in both methods, and the dashed diagonal indicates equal MAE.}
        \label{fig:peak_analysis}
    \end{minipage}
\end{figure}

To qualitatively examine the spectral decomposition capability, Fig.~\ref{fig:peak_decomposition} presents a representative example in which the reconstructed spectrum is decomposed into individual peak components. The SpectralTF-MAE reconstruction closely follows the observed spectrum, while the predicted components are consistent with the corresponding ground-truth peaks in peak position and overall line shape.

To further assess spectral fidelity quantitatively, Fig.~\ref{fig:peak_analysis} compares the peak-center localization MAE of the baseline method and SpectralTF-MAE on paired samples with the correct peak count in both methods. Most points lie below the dashed diagonal, indicating that SpectralTF-MAE achieves lower peak-center localization error for a substantial fraction of the evaluated samples. As summarized in the figure, the median MAE decreases from 0.652 to 0.575 and the mean MAE decreases from 0.955 to 0.803, demonstrating improved spectral peak localization accuracy.

\noindent $\bullet$ \textbf{Plasma-Parameter Prediction.~}

\begin{figure}[!t]
    \centering
    \subfigure[Electron-temperature ($T_e$) relative error.]{
        \includegraphics[width=0.47\textwidth]{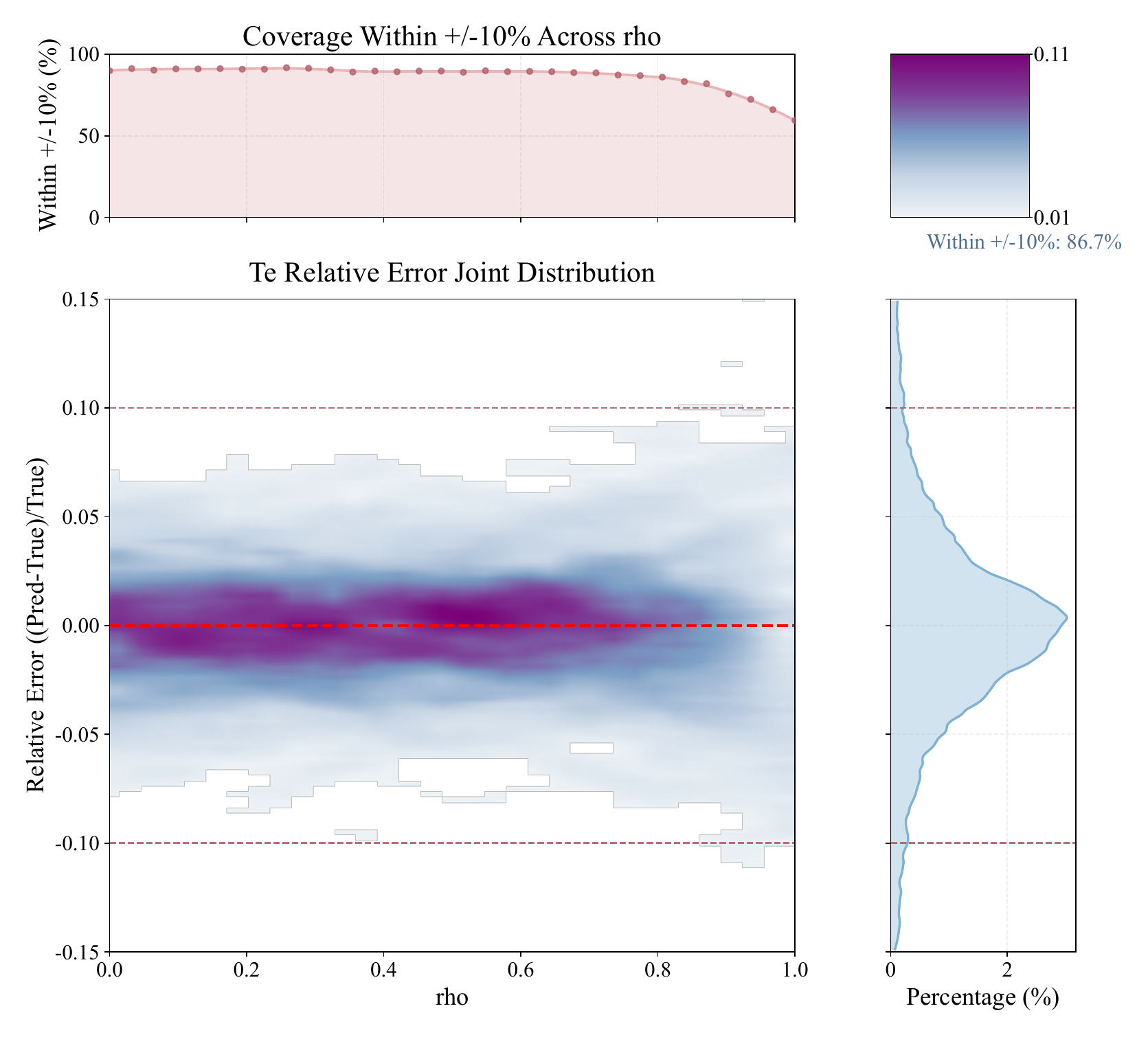}
    }
    \hfill
    \subfigure[Ion-temperature ($T_i$) relative error.]{
        \includegraphics[width=0.47\textwidth]{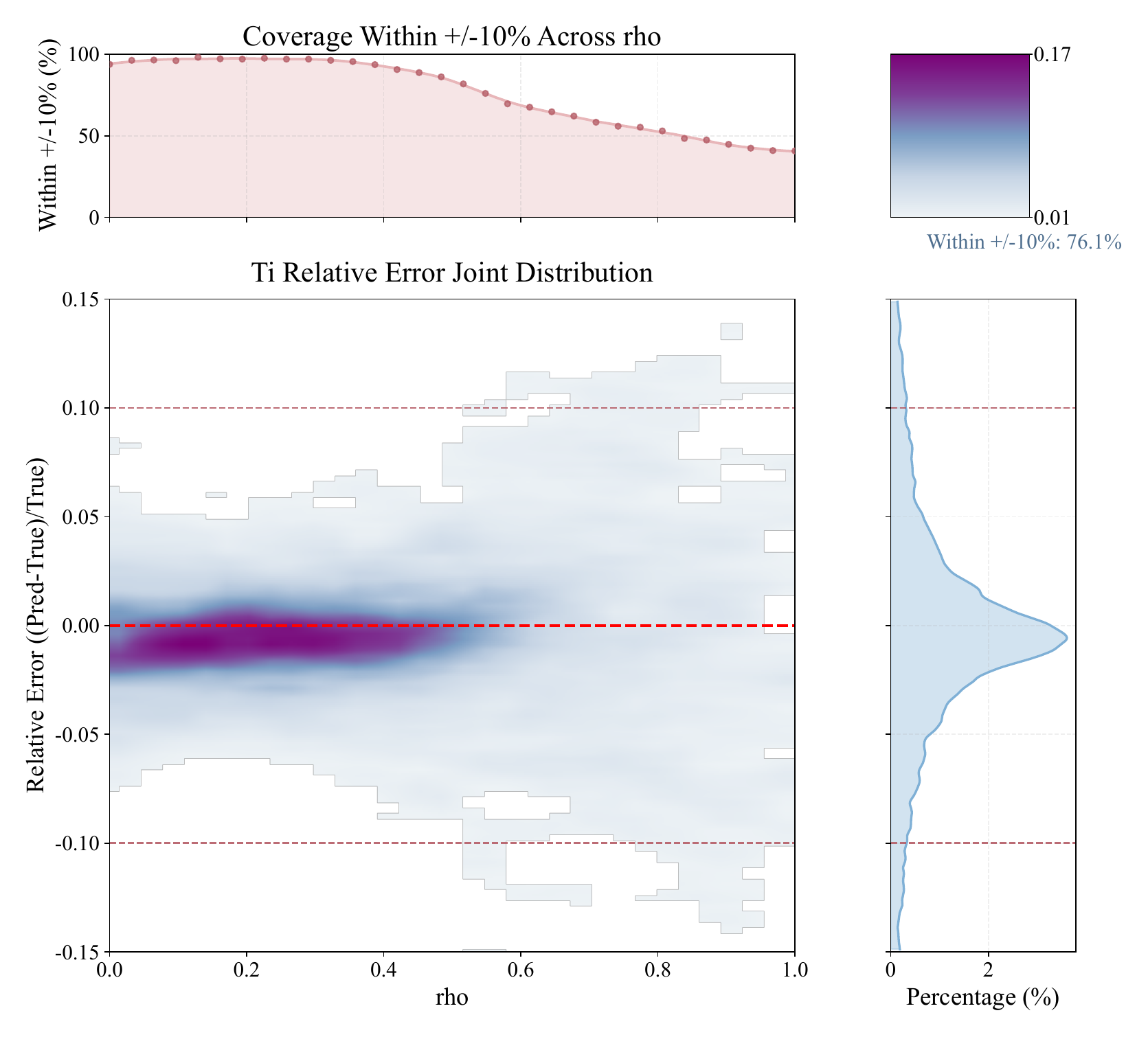}
    }
    \caption{Relative-error distributions of predicted plasma temperatures: (a) electron temperature $T_e$ and (b) ion temperature $T_i$. Each subfigure shows the error distribution as a function of the normalized radius $\rho$, together with the within-$\pm 10\%$ coverage curve and the marginal error distribution.}
    \label{fig:temperature_prediction_error}
\end{figure}

We further visualize plasma-parameter prediction from two complementary perspectives. For the temperature profiles, Fig.~\ref{fig:temperature_prediction_error}~(a) and Fig.~\ref{fig:temperature_prediction_error}~(b) show the relative-error distributions of $T_e$ and $T_i$, respectively, as functions of the normalized radius $\rho$. Both subfigures also report the radial coverage within $\pm 10\%$ error and the corresponding marginal error distribution. To avoid unstable ratios when the ground-truth value approaches zero, relative error is displayed only where the magnitude of the ground-truth value is at least 3\% of the maximum magnitude of the corresponding profile.

As shown in Fig.~\ref{fig:temperature_prediction_error}~(a) and (b), the displayed relative errors are concentrated near zero over most radial positions. The within-$\pm 10\%$ coverage is 86.7\% for $T_e$ and 76.1\% for $T_i$, while a broader spread appears toward the outer radial region, especially for $T_i$. These distributions are consistent with the overall profile-level prediction accuracy reported quantitatively.

\begin{figure}[!t]
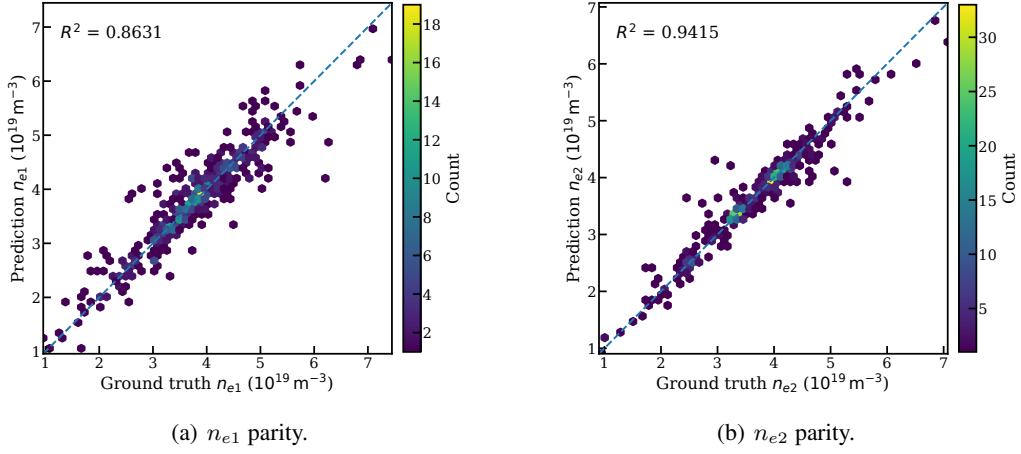

    \centering
    \subfigure[$n_{e1}$ parity.]{
        \includegraphics[width=0.47\textwidth,trim=0 0 224.55bp 16bp,clip]{figures/FigS_density_parity.pdf}
    }
    \hfill
    \subfigure[$n_{e2}$ parity.]{
        \includegraphics[width=0.47\textwidth,trim=224.55bp 0 0 16bp,clip]{figures/FigS_density_parity.pdf}
    }
    \caption{Parity plots for electron-density prediction: (a) $n_{e1}$ and (b) $n_{e2}$. The solid markers represent model predictions against the corresponding ground-truth values, and the dashed diagonal denotes ideal agreement.}
    \label{fig:ne_parity}
\end{figure}

For the electron-density quantities, Fig.~\ref{fig:ne_parity}~(a) and Fig.~\ref{fig:ne_parity}~(b) show parity plots for $n_{e1}$ and $n_{e2}$, respectively. Most predictions lie close to the identity line over the observed density range. The $n_{e2}$ predictions in Fig.~\ref{fig:ne_parity}~(b) are more tightly concentrated around the identity line, whereas $n_{e1}$ in Fig.~\ref{fig:ne_parity}~(a) exhibits greater dispersion for a subset of samples. Together with the temperature-profile results, these visualizations provide qualitative evidence that the pretrained XCS representation transfers to both scalar and profile-based plasma-parameter prediction.

\section{Limitation Analysis}\label{app:limitations}

Although SpectralTF-MAE demonstrates effective transferability across multiple plasma diagnostic tasks, several limitations remain. First, the current pretraining framework relies on paired CXRS and XCS measurements acquired under corresponding plasma states. While such paired observations facilitate the learning of cross-diagnostic correlations, their availability may constrain the amount of usable pretraining data. Extending the framework to partially paired or unpaired diagnostic measurements would further improve its applicability. Second, the present experiments are conducted using data from EAST and the corresponding diagnostic configurations. The generalization capability of the learned spectral representations across different tokamaks, diagnostic systems, and operating conditions therefore requires further validation. Finally, the peak-enhanced objective and downstream spectral-generation supervision employ predefined characteristic spectral regions. Although these regions effectively capture the dominant plasma-induced spectral responses considered in this study, they may require adjustment when the spectral calibration, impurity species, or diagnostic configuration changes. Developing adaptive mechanisms for identifying diagnostically informative spectral regions represents an important direction for future work. In addition, the Base/Large comparisons show that increasing model capacity does not uniformly improve every target or metric, suggesting that a more systematic study of architecture scaling and task-specific capacity requirements is warranted.

\end{document}